\documentclass{article}

\usepackage{PRIMEarxiv}

\usepackage[utf8]{inputenc} 
\usepackage[T1]{fontenc}    
\usepackage{hyperref}       
\usepackage{url}            
\usepackage{booktabs}       
\usepackage{amsfonts}       
\usepackage{nicefrac}       
\usepackage{microtype}      
\usepackage{lipsum}
\usepackage{fancyhdr}       
\usepackage{graphicx}       
\graphicspath{{media/}}     

\usepackage{amssymb}
\usepackage{amsmath}

\usepackage{xcolor}
\usepackage{hyperref}
\usepackage{algorithm}
\usepackage{algorithmic}
\usepackage{amsmath}
\usepackage{comment}

\title{Risk-averse policies for natural gas futures trading using distributional reinforcement learning
}

\author{
  Félicien Hêche  \\
  School of Engineering and Management \\
  University of Applied Sciences and \\
  Arts Western Switzerland (HES-SO) \\
  Yverdon-les-Bains, Switzerland \\
  \texttt{felicien.heche@gmail.com} \\
  \And
  Biagio Nigro \\
  Predictive Layer SA \\
  Morges, Switzerland \\
  \texttt{biagio.nigro@gmail.com} \\
  \And
  Oussama Barakat \\
  SINERGIES Laboratory \\
  University of Bourgogne-Franche-Comté \\
  Besançon, France \\
  \texttt{oussama.barakat@univ-fcomte.fr} \\
  \And
  Stephan Robert-Nicoud \\
  School of Engineering and Management \\
  University of Applied Sciences and \\
  Arts Western Switzerland (HES-SO) \\
  Yverdon-les-Bains, Switzerland \\
  \texttt{stephan.robert@hes-so.ch}
}

\begin{document}
\maketitle

\begin{abstract}
Financial markets have experienced significant instabilities in recent years, creating unique challenges for trading and increasing interest in risk-averse strategies. Distributional Reinforcement Learning (RL) algorithms, which model the full distribution of returns rather than just expected values, offer a promising approach to managing market uncertainty. This paper investigates this potential by studying the effectiveness of three distributional RL algorithms for natural gas futures trading and exploring their capacity to develop risk-averse policies. Specifically, we analyze the performance and behavior of Categorical Deep Q-Network (C51), Quantile Regression Deep Q-Network (QR-DQN), and Implicit Quantile Network (IQN). To the best of our knowledge, these algorithms have never been applied in a trading context. These policies are compared against five Machine Learning (ML) baselines, using a detailed dataset provided by Predictive Layer SA, a company supplying ML-based strategies for energy trading. The
main contributions of this study are as follows. (1) We demonstrate that distributional RL algorithms significantly outperform classical RL methods, with C51 achieving performance improvement of more than 32\%. (2) We show that training C51 and IQN to maximize CVaR produces risk-sensitive policies with adjustable risk aversion. Specifically, our ablation studies reveal that lower CVaR confidence levels increase risk aversion, while higher levels
decrease it, offering flexible risk management options. In contrast, QR-DQN shows less predictable behavior. These findings emphasize the potential of distributional RL for developing adaptable, risk-averse trading strategies in volatile markets.
\end{abstract}

\section{Introduction}\label{intro}

Financial markets have experienced significant instabilities in recent years, attributable to several factors. One notable source of these instabilities is the COVID-19 pandemic, which has affected many financial assets \cite{ali2020coronavirus, zhang2020financial}. For example, given that COVID-19 has significantly impacted energy demand and consumption \cite{aruga2020effects, jiang2021impacts}, it is unsurprising that this pandemic has increased volatility in energy markets \cite{huang2023time}. 
Another major source of instability is the ongoing conflict between Ukraine and Russia, which has substantially heightened volatility across various financial assets \cite{fang2022russia, lin2024impact, ngwakwe2022stock,  rice2023russia, wu2023stock}, particularly affecting markets reliant on Russian commodities \cite{lo2022russo}. Specifically, the strategic use of Russian gas supply as a geopolitical instrument in the tension opposing Russia and Europe \cite{virag2023turbulent, zhang2023tensions} coupled with the fact that Russia was one of the biggest gas suppliers to European Union countries a few years ago \cite{report2022}, has precipitated pronounced instabilities in the European gas market and derived financial products \cite{enescu2023discussing, inacio2023assessing, report2022}. These fluctuations pose specific challenges for trading and increase the interest in risk-averse strategies.

In light of the notable advancements provided by Machine Learning (ML) algorithms over the last decade \cite{brown2020language, he2016deep, simonyan2014very}, a growing body of research has focused on leveraging these algorithms to develop new trading strategies \cite{ayitey2023forex, huang2019automated, lv2019empirical}. Specifically, Reinforcement Learning (RL), a ML paradigm, has been extensively employed in the development of innovative trading methods \cite{deng2016deep, shavandi2022multi, wu2020adaptive}.

Within the RL framework, an agent interacts with an environment that produces rewards in each time step \cite{sutton2018reinforcement}. Classical RL policies aim to maximize the expectation of the cumulative rewards.
A specific class of RL algorithms, known as distributional RL, explicitly models the entire distribution of the cumulative rewards, rather than solely focusing on their expected value \cite{bellemare2023distributional}. Distributional RL has shown impressive results over the past years \cite{bellemare2017distributional, dabney2018implicit, dabney2018distributional}, including in challenging stochastic environment \cite{heche2024offline}. Furthermore, distributional RL enables the construction of policies that maximize a risk measure induced by the cumulative rewards. Throughout this paper, such policies will be referred to as risk-sensitive policies, contrasting with risk-free policies that aim to maximize the expected return value.

Previous research suggests that training policies with a risk measure called Conditional Value-at-Risk (CVaR) \cite{artzner1999coherent, sarykalin2008value} offer several notable properties. For instance, some works highlight that training policies with CVaR can yield robustness in terms of differences between training and test conditions \cite{pinto2017robust} or system disturbance \cite{singh2020improving}. Additionally, training policies considering CVaR might also build risk-averse policies
\cite{ma2021conservative, rigter2021risk, urpi2021risk}.

These studies suggest that distributional RL holds interesting properties to trade challenging and volatile assets, such as natural gas futures. Futures are standardized financial contracts obligating the buyer to purchase, and the seller to sell, a specific asset at a predetermined price on a future date \cite{chance2021introduction}. The notable performance observed in demanding stochastic environments may translate effectively to highly volatile markets. Subsequently, distributional RL algorithms may offer the availability to construct risk-averse policies, that appear to be particularly valuable for trading in unstable markets. However, the performance of these policies and their ability to develop risk-averse behavior remain uncertain in this context. 
Furthermore, it is unclear whether the risk aversion of these risk-sensitive policies can be finely adjusted using the confidence level of CVaR. 

This paper aims to investigate the effectiveness of three distributional RL algorithms for natural gas futures trading and explore their potential to develop risk-averse policies. Specifically, we analyze the performance and behavior
of Categorical Deep Q-Network (C51) \cite{bellemare2017distributional}, Quantile Regression Deep Q-Network (QR-DQN) \cite{dabney2018distributional}, and Implicit Quantile Network (IQN) \cite{dabney2018implicit}. To the best of our knowledge, these distributional RL algorithms have never been applied in a trading context. These policies are compared against five baselines, including four classical RL agents and one ML model supplied by Predictive Layer SA, a company providing ML-based strategies for energy trading. Four experiments are conducted using a meticulously constructed and preprocessed historical dataset supplied by the same company. This dataset spans from the beginning of 2017 to the end of 2022 and contains 1141 features associated with each specific futures and trading day. The
main contributions of this work are as follows. (1) We demonstrate that distributional RL algorithms significantly outperform classical RL methods, with C51 achieving a performance improvement of more than 32\%. (2) We show that training C51 and IQN to maximize CVaR produces risk-sensitive policies with adjustable risk aversion. Specifically, our ablation studies reveal that lower CVaR confidence levels increase risk aversion, while higher levels decrease it, offering flexible risk management options. In contrast, QR-DQN
shows less predictable behavior. These findings emphasize the potential of distributional RL for developing adaptable, risk-averse trading strategies in volatile markets.

This paper is organized as follows. First, Section \ref{related work} provides a comprehensive literature review. Section \ref{preliminaries} revisits fundamental concepts and introduces the notation adopted throughout this paper. Then, Section \ref{RL agents} describes trading agents employed in this study. Experimental setup and results are fully detailed in Section \ref{Experiments}. Following this, Section \ref{Discussion} delves into a discussion of the findings derived from these experiments. Finally, the concluding remarks are presented in the last section.

\section{Related work}\label{related work}

\subsection{Machine learning in finance}

Over the past decade, ML has seen widespread application across various domains within finance \cite{ahmed2022artificial, levchenko2024chain, nazareth2023financial}. For example, some studies utilize ML techniques for bankruptcy prediction
\cite{jones2017predicting, lahmiri2019can}, while others focus on detecting money laundering \cite{garcia2021ai, jullum2020detecting} or fraud detection \cite{omar2017predicting}.

\subsection{Machine learning for trading}

ML has been extensively applied in financial trading, with models primarily designed for market trend prediction \cite{gorenc2016prediction, lv2019empirical} or asset price forecasting \cite{bao2017deep, chalvatzis2020high, yong2017stock}. A wide range of ML models, including Random Forest \cite{breiman2001random} and Long Short-Term Memory Networks (LSTM) \cite{hochreiter1997long}, have been employed for these tasks \cite{ghosh2022forecasting}. However developing effective ML models for trading remains challenging due to the high volatility and non-stationary nature of financial markets, which are further influenced by unpredictable external factors such as economic policy changes and geopolitical events.

To address this inherent uncertainty, researchers have explored strategies to optimize the portfolio Sharpe Ratio \cite{zhang2020deep} and have developed quantile regression approaches \cite{zhang2019extending}. Another promising method for managing market uncertainties involves using neural networks to model the probability distribution of asset prices \cite{marcjasz2023distributional}.

\subsection{Reinforcement learning in finance}

RL has also found widespread application in finance \cite{hambly2023recent}. For instance, numerous researchers have proposed RL methods to tackle the optimal execution problem  \cite{lin2021end, ye2020optimal}. This problem entails determining the most efficient way to execute a large trade while minimizing costs such as market impact and price slippage. One of the most popular RL algorithms used in this domain is Deep Q-Network (DQN) \cite{ning2021double}. Additionally, RL has been applied to options pricing and hedging  \cite{cao2021deep, du2020deep, halperin2017qlbs}. Option pricing involves determining the fair value of a financial derivative contract, while hedging entails managing risk by taking offsetting positions.

 \subsection{Reinforcement learning for trading}
One of the most intensive applications of RL in finance involves trading and portfolio optimization  \cite{aboussalah2022value, cong2021alphaportfolio, pendharkar2018trading}. However, these applications primarily rely on classical RL methods like DQN, which stands out as the most widely used RL algorithm for trading 
\cite{brim2020deep, chen2019application, park2020intelligent, sornmayura2019robust, theate2021application, tran2023optimizing}. Other studies also incorporate classical RL policies, such as Deep Deterministic Policy Gradient \cite{liu2018practical}, or Dueling DQN \cite{li2020application}. Additionally, various works propose enhancements to improve the performance of these trading agents. For instance, \cite{kwak2023self} incorporates a self-attention mechanism into the training process, while another study introduces a data augmentation-based RL framework to enable training RL policies with limited data \cite{yuan2020using}. Moreover, ensemble RL \cite{karkhanis2020ensembling} or hierarchical RL \cite{shavandi2022multi} approaches have also been explored.

Despite the potential of distributional RL to address the inherent uncertainty of financial markets, its application in this domain remains surprisingly limited.  Nevertheless, it is worth mentioning that this class of algorithms has been applied in financial contexts outside of trading \cite{cao2023gamma, fathan2021deep, vittori2020option}.

Distributional RL allows training agents to minimize a measure of the risk induced by the cumulative rewards, instead of maximizing the expected return. This framework has gained attention in recent years beyond financial applications
\cite{fei2021exponential, greenberg2022efficient, theate2023risk}. Various risk measures have been considered, such as Cumulative Prospect Theory \cite{tversky1992advances} or CVaR. CVaR is popular and has been studied in the context of RL for many years \cite{chow2014algorithms, chow2015risk, ying2022towards}.

While some studies mention building risk-averse trading RL policies as future work \cite{hambly2023recent, zhang2019deep}, surprisingly little research has been conducted in this area. However, it is noteworthy that \cite{wang2022risk} integrates distributional RL into a hierarchical RL framework. Additionally, some authors enhance the classical Q-learning algorithm to build risk-averse RL trading agents \cite{shen2014risk}. Nevertheless, it is important to note that these methods, which rely on the Q-learning algorithm, may yield promising outcomes solely in scenarios characterized by discrete and relatively small state spaces, unlike the algorithms studied in this paper.

\section{Preliminaries}\label{preliminaries}

Before delving deeper into our research, we will review essential concepts and establish the notation employed throughout this paper. First, we will provide a concise introduction to CVaR. Following that, we will recall basic RL principles.

\subsection{Conditional Value-at-Risk} 
Let $X$ denote a bounded-mean random variable on a probability space  $(\Omega, \mathcal{F}, \mathbb{P})$ and $F$, defined as $F(x) :=   \mathbb{P}(X \leq x) $, its cumulative distribution function. In this paper, $X$ should be interpreted as the cumulative rewards or any variable we aim to maximize. Value-at-Risk (VaR)  at confidence level $\alpha \in (0, 1)$ is defined as $\text{VaR}_{\alpha}(X) := \text{inf}\{ x : F(x) \geq \alpha \}$. In other words, VaR$_{\alpha}(X)$ is the $\alpha$-quantile of $X$. Using VaR, it is possible to define CVaR \cite{acerbi2002spectral}
\begin{equation}\label{aebi formula}
    \text{CVaR}_{\alpha}(X) := \frac{1}{\alpha} \int_{0}^{\alpha} \text{VaR}_{\gamma}(X) d\gamma
\end{equation}
where $\alpha$ is called the confidence level. More intuitively, and if $X$ has a continuous distribution, we have
\begin{equation}
    \text{CVaR}_{\alpha}(X) = \mathbb{E}\left[ X | X \leq \text{VaR}_{\alpha}(X) \right].
\end{equation}
Thus, CVaR might be interpreted as the expected value of $X$ under the $\alpha$ worst-case scenario. Remark that if the confidence level tends to one, CVaR tends to the expectation of $X$.

Furthermore, since CVaR is a coherent risk measure \cite{rockafellar2007coherent}, it satisfies several properties. One of these states that each coherent risk measure $\mathcal{R}$ admits a risk envelope $\mathcal{U}$ such that 
\begin{equation}\label{dual representation}
\mathcal{R}(X) = \inf_{\delta \in \mathcal{U}} \mathbb{E} \left[ \delta X \right]
\end{equation}
\cite{artzner1999coherent, delbaen2002coherent, rockafellar2002deviation}. A risk envelope is a closed nonempty convex subset of $ \mathcal{P} := \{ \delta \in \mathcal{L}^{2} \text{ } :  \text{ } \text{ } \delta \geq 0 \text{, } \int_{\Omega} \delta(\omega) d\mathbb{P}(\omega) = 1 \}$. 
Moreover, the risk envelope associated with CVaR, can be written as  
\begin{equation} \label{dual cvar}
\mathcal{U} = \left\{ \delta  \in \mathcal{P} \text{ } : \text{ }\delta(\omega) \in \left[0, \frac{1}{\alpha}\right]\text{, } \int_{\Omega} \delta(\omega) d\mathbb{P}(\omega) = 1  \right\}
\end{equation}
\cite{rockafellar2002deviation, rockafellar2006generalized}. The formulation of CVaR$_{\alpha}(X)$ using the risk envelope, known as the dual representation, underscores the robust nature of CVaR. Essentially,  CVaR may be interpreted as the expectation of $X$ under a
worst-case perturbed distribution. Notably, the definition of the risk envelope guarantees the validity of the perturbed distribution, while enabling an increase in the probability density of any outcome by a factor of at most $1/\alpha$.

\subsection{Reinforcement learning}

Following the classical RL framework, we adopt a Markov Decision Process (MDP) to model agent-environment interactions. An MDP is represented as a tuple $(S, \mathcal{A}, P, R, \mu_{0}, \gamma)$, where $S$ denotes the environment space and $\mathcal{A}$ the action space. The reward function $R(s_{t}, a_{t})$ is treated as a random variable in this study. Rewards obtained at time $t$ will be referred to as $r_{t}$. Additionally, $P$ is the transition probability distribution ($s_{t+1} \sim P( \cdot | s_{t}, a_{t})$), $\mu_{0}$ represents the initial state distribution, and $\gamma \in (0, 1)$ denotes the discount factor. For the purpose of the notation, we define $P(s_{0}) := \mu_{0}(s_{0})$, and denote the MDP as $(S, \mathcal{A}, P, R, \gamma)$.

In RL, actions are dictated by a policy $\pi$ which depends on the environment state $s_{t}$.  We use the notation $a_{t} \sim \pi$ to denote that the action $a_{t}$ has been chosen according to the policy $\pi$. A sequence $\tau = s_{0}, a_{0}, r_{0},  \ldots s_{H}$, with $s_{i} \in S$ and $a_{i} \in \mathcal{A}$ is called a trajectory. An episode refers to a trajectory of fixed length $H \in \mathbb{N}$. Given, a policy $\pi$ and a sate-action couple $(s_{t}, a_{t})$, the associated Q-function is defined as
\begin{equation}
   Q_{\pi}(s_{t}, a_{t}):= \mathbb{E}_{\pi} \left[\sum_{t'=t}^{H} \gamma^{t'-t} R(s_{t'}, a_{t'}) \text{ } \big \rvert \text{ } s_{t}, a_{t} \right]. 
\end{equation}

The goal of classical risk-free RL algorithms is to construct a policy $\pi$ that maximizes the expected discounted return $\mathbb{E}_{\pi} \left[ \sum_{t=0}^{H} \gamma^{t}R(s_{t}, a_{t}) \right]$ where $H$ may be infinite. The Q-function associated with the optimal policy $\pi_{\ast}$ is denoted $Q_{\ast}$. The optimal Q-function satisfies the following equality known as the Bellman equation
\begin{equation}
    Q_{\ast}(s_{t}, a_{t}) = \mathbb{E}\left[ R(s_{t}, a_{t}) \right] + \gamma \mathbb{E} \left[ \text{max}_{a \in \mathcal{A}} \text{ } Q_{\ast}(s_{t+1}, a)\right]
\end{equation}
where $s_{t+1} \sim P(\cdot | s_{t}, a_{t})$.

In risk-sensitive RL, the objective is to devise policies that maximize a given risk-measure $\mathcal{R}$ applied to the cumulative discounted reward, expressed as $\mathcal{R}\left(\sum_{t=0}^{H} \gamma^{t}R(s_{t}, a_{t}) \right)$.

\section{Trading agents}\label{RL agents}
This section aims to offer an overview of all agents employed in this study. We begin by introducing the ML trading model, followed by a presentation of the four risk-free RL policies. Finally, concise descriptions of the three distributional RL algorithms utilized in this study will be provided.

\subsection{Extra-Trees}
Our study includes an ML model provided by Predictive Layer SA: an Extra-Trees model \cite{geurts2006extremely}. Extra-Trees is an ensemble of trees designed to enhance Random Forest. Extra-trees differs from Random Forest in two key aspects
: it utilizes the entire training dataset to grow the trees and selects cut points for node splitting fully at random.
This randomized node-splitting process yields a more diverse set of trees, resulting in a more robust model.  Empirical results have indicated superior predictive performances of this model over Random Forest in forecasting financial market fluctuations \cite{pagliaro2023forecasting}.

In our experiments, the Extra-Trees model has been trained as a binary classifier, forecasting market direction, i.e., whether the market will rise or fall. Throughout our experiments, we utilize an Extra-Trees model with hyperparameters supplied by Predictive Layer SA.

\subsection{Classical RL}
Now, we present all four classical RL algorithms used in this study. These algorithms encompass DQN and three of its extensions.

\subsubsection{DQN}
Our DQN implementation follows \cite{mnih2015human} with minor adjustments. Specifically, we use the double Q-learning method \cite{van2016deep} and the clip trick \cite{fujimoto2018addressing}.

In other words, we use two neural networks $Q_{\theta_{i}} \text{ : } S \times \mathcal{A} \rightarrow \mathbb{R}$ with $i \in \{ 1, 2\}$ to approximate $Q_{\ast}$. As a common notation, $\theta_{i}$ denotes the weight of the neural network $Q_{\theta_{i}}$. The associated policy consists of taking action that maximizes the mean of the two Q-models
\begin{equation}
    \pi_{DQN}(s_{t}) = \text{argmax}_{a \in \mathcal{A}} \text{ } \frac{1}{2} \left( Q_{\theta_{1}}(s_{t}, a)  + Q_{\theta_{2}}(s_{t}, a)\right).
\end{equation}
To train $Q_{\theta_{i}}$, DQN alternates between environment and learning steps. During the environment steps, the agent interacts with the environment to collect transitions of the form $(s_{t}, a_{t}, r_{t}, s_{t+1})$, where $a_{t} \sim \pi_{DQN}$. These tuples are then stored in a replay buffer denoted $\mathcal{B}$. Subsequently, during the learning step, gradient descent is performed on $Q_{\theta_{i}}$ to minimize the mean square error between $\mathcal{T}Q_{\overline{\theta}_{i}}$ and $Q_{\theta_{i}}$, also called Temporal Difference (TD) error. The operator $\mathcal{T}$, called the Bellman operator,  is defined as
\begin{equation}
    \mathcal{T}Q_{\overline{\theta}_{i}}(s_{t}, a_{t}) = 
    r_{t} + \gamma \min_{j}Q_{\overline{\theta}_{j}}\left(s_{t+1}, \pi_{DQN}(s_{t+1}) \right)
\end{equation}
and $Q_{\overline{\theta}_{j}}$ is called the target network. At the outset of the training, $Q_{\overline{\theta}_{j}}$ is initialized as an exact copy of $Q_{\theta_{j}}$. Subsequently, at each training step, $\overline{\theta}_{j}$ undergoes a smooth update using $\overline{\theta}_{j} \leftarrow \tau \overline{\theta}_{j} + (1 - \tau)\theta_{j}$, where $\tau$ represents a hyperparameter typically set to a value close to one. Thus, using these notations, $Q_{\theta_{i}}$ is trained to minimize
\begin{equation}
    \mathbb{E}_{(s_{t}, a_{t}, r_{t}, s_{t+1}) \sim \mathcal{B}} \left[ 
    \left( \mathcal{T}Q_{\overline{\theta}_{i}}(s_{t}, a_{t})  - Q_{\theta_{i}}(s_{t}, a_{t})
    \right)^{2}
    \right].
\end{equation}
The $Q_{\theta_{i}}$ models used in this study are composed of one LSTM layer with $128$ units, followed by two dense layers with $64$ and $32$ units, respectively. These neural networks have been trained using a learning rate of $1 \times 10^{-4}$, and a batch size of $32$. Similarly to all neural networks used in this paper, these models incorporate batch normalization \cite{ioffe2015batch} and dropout \cite{srivastava2014dropout}, with a rate of $0.3$. Furthermore, they use the Rectified Linear Unit (ReLU) as activation function and have been trained using Adam optimizer \cite{kingma2014adam}.

\subsubsection{Prioritized DQN}

Our first extension of DQN, called Prioritized DQN, consists of adding a prioritized replay \cite{schaul2015prioritized} into the DQN framework. In Prioritized DQN, unlike in DQN, transitions $(s_{t}, a_{t}, r_{t}, s_{t+1})$ are not uniformly randomly drawn from the replay buffer. Instead, they are sampled from the replay buffer $\mathcal{B}$ based on a probability $p_{t}$ that is proportional to mean of the TD error
\begin{equation}
    p_{t} \propto \frac{1}{2} \left(  \left( \mathcal{T}Q_{\overline{\theta}_{1}}(s_{t}, a_{t}) - Q_{\theta_{1}}(s_{t}, a_{t}) \right)^{2}
    +
    \left( \mathcal{T}Q_{\overline{\theta}_{2}}(s_{t}, a_{t}) - Q_{\theta_{2}}(s_{t}, a_{t})\right)^{2}
    \right).
\end{equation}
To enable this prioritization, the replay buffer is implemented using a sum tree data structure, affording a $O(\log(M))$ complexity for both sampling and adding transitions, where $M$ denotes the number of transitions stored into $\mathcal{B}$.

The $Q_{\theta_{i}}$ models used in Prioritized DQN are composed of one LSTM with $128$ units, followed by two fully connected dense layers with $32$ and $16$ units, respectively. The training process involves a learning rate of $5 \times 10^{-5}$, and a batch size of $128$.

\subsubsection{Dueling DQN}

\begin{figure}
	\centering
		\includegraphics[width = 6cm]{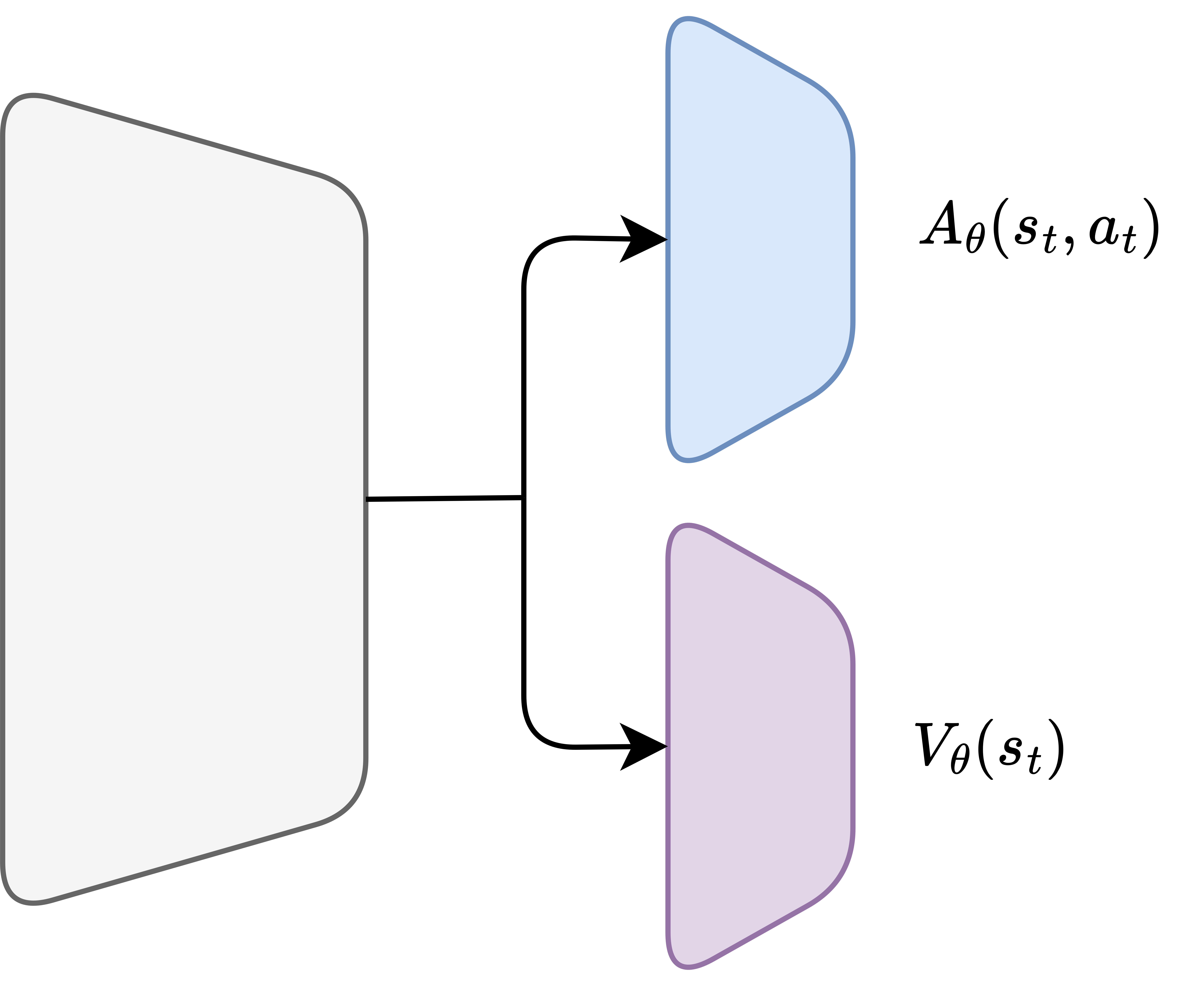}
	\caption{Illustration of the neural architecture used in Dueling DQN.}
	\label{dueling architecture}
\end{figure}

Dueling DQN is an extension of DQN that aims to improve its performance using a specific dueling network architecture \cite{wang2016dueling}. As illustrated in Figure \ref{dueling architecture}, this network comprises two models designated as $V_{\theta}$ and $A_{\theta}$, which share a common component. These models aim to approximate the value function $V_{\pi}(s_{t}) = \mathbb{E}_{a \sim \pi}\left[ Q_{\pi}(s_{t}, a) \text{ } |\text{ } s_{t} \right]$ and the advantage function $A_{\pi}$ defined as $A_{\pi}(s_{t}, a_{t}):= Q_{\pi}(s_{t}, a_{t}) - V_{\pi}(s_{t})$, respectively.
Furthermore, we introduce two pairs models $V_{\theta_{1}}, V_{\theta_{2}}, A_{\theta_{1}}, A_{\theta_{2}}$ and their corresponding target networks. These models enable to estimate the optimal Q-function using the following equation 
\begin{equation} \label{dueling q}
Q_{\theta_{i}}(s_{t}, a_{t}) = V_{\theta_{i}}(s_{t})
+ \left( A_{\theta_{i}}(s_{t}, a_{t}) - \frac{1}{| \mathcal{A} |} \sum_{a'} A_{\theta_{i}}(s_{t}, a')
\right).
\end{equation}
Then, Dueling DQN follows the method proposed in DQN to train these models. The subtraction of the mean of the advantage function in the above equation might initially appear unexpected. However, this adjustment is crucial to maintain identifiability. Without it, the equation becomes unidentifiable, meaning that given $Q_{\theta_{i}}(s_{t}, a_{t})$, $V_{\theta_{i}}(s_{t})$ and $A_{\theta_{i}}(s_{t}, a_{t})$ can not be recovered uniquely, resulting in poor practical performance.

In our experiments, the shared component of the dueling architecture comprises two stacked LSTM with $128$ and $64$ units, followed by a fully dense layer consisting of $64$ units. Subsequently, the components specifically dedicated to $A_{\theta_{i}}, V_{\theta_{i}}$ consist both of one dense layer with $64$ units. These models have been trained using a batch size of $64$ and a learning rate of $5\times 10^{-5}$.

\subsubsection{Prioritized Dueling DQN}
The last classical RL algorithm used in this study integrates both enhancements proposed in Dueling and Prioritized DQN. Specifically, this policy, called Prioritized Dueling DQN, is a Dueling DQN that uses a prioritized replay as introduced above.

The Prioritized Dueling model used in this study matches the architecture used with Dueling DQN, with the sole adjustment being the modification of the batch size to $16$.

\subsection{Distributional RL}
In this subsection, we introduce the distributional policies under investigation, C51, QR-DQN, and IQN.

\subsubsection{C51}
In C51 \cite{bellemare2017distributional}, rather than solely focusing on learning the expected cumulative return, the objective is to learn the distributions of the random variable  $Z(s_{t}, a_{t}) = \sum_{t' \geq t} \gamma^{t'}R(s_{t'}, a_{t'})$. For this purpose, given a policy $\pi$, a distributional variation of the Bellman operator denoted $\mathcal{\hat{T}}$, is introduced
\begin{equation}
    \mathcal{\hat{T}}Z(s_{t}, a_{t}) \overset{\text{D}}{=}  R(s_{t}, a_{t}) +\gamma  Z(s_{t+1}, a_{t+1})
\end{equation}
where, $s_{t+1} \sim P(\cdot | s_{t}, a_{t})$, $a_{t+1} \sim \pi$, and $\text{ }\overset{\text{D}}{=} \text{ }$ denotes an equality in distribution. Next, two fixed values $v_{min}, v_{max}$ along with a predetermined number of atoms $N_{atoms}$ are established. Following common practice, we set $N_{atoms}$ to $51$ in our experiments. Moreover, the set of atoms $\{ z_{j} \text{ } : \text{ } 0 \leq  j < N_{atoms} \}$ where $z_{j}$ is defined as
\begin{equation}
z_{j} := v_{min} + j \frac{v_{max} - v_{min}}{N_{atoms}-1} = v_{min} + j \Delta z
\end{equation}
is introduced. This set of atoms serves the purpose of discretizing the possible values that $Z$ could take on.

Furthermore, C51 introduces a model $P_{\theta}$, whose produces outputs of dimension $N_{atoms}$ aiming to approximate the probability distribution of $Z(s_{t}, a_{t})$. Specifically, we have $P_{\theta}(s_{t}, a_{t})_{j} \simeq \mathbb{P}\left(  Z(s_{t}, a_{t}) = z_{j})\right)$. The associated policy $\pi_{C51}$ is defined as 
\begin{equation}
    \pi_{C51}(s_{t}) := \text{argmax}_{a \in \mathcal{A}} \sum_{i} z_{i}P_{\theta}(s_{t}, a)_{i} \simeq \text{argmax}_{a \in \mathcal{A}} \text{ } Q_{\pi_{C51}}(s_{t}, a_{t}).
\end{equation}
Similarly to DQN, C51 alternates between environment steps, where the agent interacts with the environment to add transitions into the replay buffer, and training steps. During the training steps the model $P_{\theta}$ is trained to minimize the cross-entropy between $\mathcal{\hat{T}}P_{\overline{\theta}}$ and $P_{\theta}$ where $P_{\overline{\theta}}$ denotes the target model associated to $P_{\theta}$. Following \cite{bellemare2017distributional}, this loss function is computed using the procedure described in Algorithm \ref{C51}.

In our experiments, $P_{\theta}$ consists of one LSTM layer with $128$ units followed by two fully connected layers of both $64$ units. The final activation function of our model naturally consists of a softmax. This model has been trained using a learning rate of $5 \times 10^{-5}$ and a batch size of $32$.

Furthermore, remark that the model $P_{\theta}$ allows to estimate CVaR$_{\alpha}(Z(s_{t}, a_{t}))$ using 
\begin{equation}
    \text{CVaR}_{\alpha}(Z(s_{t}, a_{t})) \simeq \frac{1}{\alpha}\sum_{z_{i} \leq q_{\alpha}} z_{i}P_{\theta}(s_{t}, a_{t})_{i}
\end{equation}
where $q_{\alpha} = \text{VaR}_{\alpha}\left( Z(s_{t}, a_{t}) \right) 
\simeq \min \{ z_{j} \text{ : } \sum_{k \leq j} P_{\theta}(s_{t}, a_{t})_{k} \geq \alpha \}$. Therefore, C51 enables building risk-sensitive policies denoted C51$_{\alpha}$ that choose actions to maximize CVaR instead of the expectation
\begin{equation}
    \pi_{C51_{\alpha}}(s_{t}) := \text{argmax}_{a \in \mathcal{A}}
    \text{CVaR}_{\alpha}\left(Z(s_{t}, a) \right).
\end{equation}
In this paper, each distributional RL algorithm has been employed to construct multiple risk-sensitive agents by adjusting the confidence level $\alpha$. Specifically, confidence levels ranging from $0.1$ to $0.9$ with an increment of $0.2$ have been tested. All risk-sensitive RL model hyperparameters have been fine-tuned using a confidence value of $0.7$. Other risk-sensitive policies use the same hyperparameters.

Models used by C51$_{\alpha}$ involve two stacked LSTM layers with both $128$ units followed by two fully connected layers of $128$ and $64$ units. These models have been trained using a batch size of $8$ and a learning rate of $1 \times 10^{-4}$.

\begin{algorithm}[H]
\begin{algorithmic}[1]
\STATE \textbf{input:} A transition $(s_{t}, a_{t}, r_{t}, s_{t+1})$
\STATE $Q_{\pi_{C51}}(s_{t+1}, a) := \sum_{i} z_{i} P_{\theta}(s_{t+1}, a)_{i}$
\STATE $a^{\ast} \leftarrow argmax_{a} Q_{\pi_{C51}}(s_{t+1}, a)$
\STATE $m_{i} = 0$, $i \in \{ 0, \ldots N_{atoms}-1 \}$.
\FOR{$j=1$ to $N_{atoms}-1$}
\STATE \# Compute the projection of $\mathcal{\hat{T}}z_{j}$ onto the support $z_{j}$
\STATE $\mathcal{\hat{T}}z_{j} \leftarrow [r_{t} + \gamma z_{j}]_{v_{min}}^{v_{max}}$
\STATE $b_{j} \leftarrow (\mathcal{\hat{T}}z_{j} - v_{min}) / \Delta z$ \quad \# $b_{j} \in [0, N_{atoms}-1 ]$
\STATE $l \leftarrow \lfloor b_{j} \rfloor$, $u \leftarrow \lceil b_{j} \rceil$
\STATE \# Distribute the probability of $\mathcal{\hat{T}}z_{j}$.
\STATE $m_{l} \leftarrow m_{l} + P_{\overline{\theta}}(s_{t+1}, a^{\ast})_{j}(b_{j} - l)$
\STATE $m_{u} \leftarrow m_{u} + P_{\overline{\theta}}(s_{t+1}, a^{\ast})_{j}(u - b_{j})$
\ENDFOR
\STATE $- \sum_{i} m_{i} \log \left( P_{\theta}(s_{t}, a_{t})_{i} \right)$ \# Cross-entropy loss
\end{algorithmic}
\caption{C51: computing cross-entropy loss}
\label{C51}
\end{algorithm}
\noindent

\subsubsection{QR-DQN}
One of the limitations of C51 lies in the fact that the set of atoms is fixed and requires manual specification. However, the distribution $Z(s_{t}, a_{t})$ may vary depending on the state-action couple, making the use of a fixed support potentially too restrictive. To overcome this limitation, QR-DQN \cite{dabney2018distributional} proposes an alternative approach by considering $L$ fixed probabilities $p_{i}$, each with a weight of $1/L$ and trying to estimate the associated quantile $q_{i}$ such that $\mathbb{P}(Z(s_{t}, a_{t}) \leq q_{i}) = \tau_{i} = \frac{i}{L}$, for $1 \leq i \leq L$. These quantiles are estimated using a model denoted $T_{\theta}$. The training of this model involves the use of the asymmetric Huber loss
\begin{equation}
    \rho_{\tau}^{k}(u)
    = | \tau - \delta_{\{ u < 0 \} }| \frac{\mathcal{L}_{k}(u)}{k}
\end{equation}
where $\mathcal{L}_{k}$ is the Huber loss \cite{huber1992robust}, defined as
\begin{equation}
    \mathcal{L}_{k}(u) = 
     \begin{cases}
      \frac{1}{2}u^{2} & \text{if }  | u | \leq k \\
      k \left( |u| - k/2 \right)        & \text{otherwise.} \\
    \end{cases}
\end{equation}
More precisely, $T_{\theta}$ is trained to minimize the following loss function
\begin{equation}
    \sum_{i=1}^{L} \mathbb{E}_{j} \big[ 
    \rho_{\hat{\tau}_{i}}^{k} \left( r_{t} + \gamma T_{\overline{\theta}}(s_{t+1}, a^{\ast})_{j}
    - T_{\theta}(s_{t}, a_{t})_{i}
    \right)
    \big]
\end{equation}
where $\hat{\tau}_{i} = \frac{\tau_{i-1} + \tau_{i}}{2}$, and the action $a^{\ast}$ has been chosen according to the policy associated with QR-DQN
\begin{align}
    a^{\ast} & =  \pi_{QR-DQN}(s_{t+1}) =  \text{argmax}_{a \in \mathcal{A}} \sum_{k} p_{k} T_{\theta}(s_{t+1}, a)_{k} \\ 
    & \simeq \text{argmax}_{a \in \mathcal{A}} \text{ }Q_{\pi_{QR-DQN}}(s_{t+1}, a).
\end{align}
In our experiments, $T_{\theta}$ is composed of two stacked LSTM layers with both $128$ units, followed by two fully connected dense layers of $64$ units. A batch size of $16$ and a learning rate of $1 \times 10^{-4}$ have been used to train this model.

Furthermore, using QR-DQN, it is possible to generate risk-sensitive policies QR-DQN$_{\alpha}$  that choose actions to maximize CVaR$_{\alpha}(Z(s_{t}, a_{t}))$. Specifically, actions are chosen according to the following policy
\begin{align}
    \pi_{QR-DQN_{\alpha}}(s_{t}) & = \text{argmax}_{a \in \mathcal{A}} \text{ CVaR}_{\alpha}(Z(s_{t}, a))  \\
    & = \text{argmax}_{a \in \mathcal{A}} \text{ }
     \frac{1}{\alpha} \int_{0}^{\alpha} \text{VaR}_{\gamma}(Z(s_{t}, a_{t})) d\gamma \\
     & \simeq \text{argmax}_{a \in \mathcal{A}} \text{ } \frac{1}{\alpha L} \sum_{k \leq \alpha L} T_{\theta}(s_{t}, a)_{k}.
\end{align}

Models that compose QR-DQN$_{\alpha}$, consist of two stacked LSTM layers both with $128$ units, followed by two fully connected dense layers with $64$ and $32$ units respectively. These models have been trained using a learning rate of $1\times 10^{-4}$ and a batch size of $64$.

\subsubsection{IQN}
\begin{figure}
	\centering
		\includegraphics[width = 10cm]{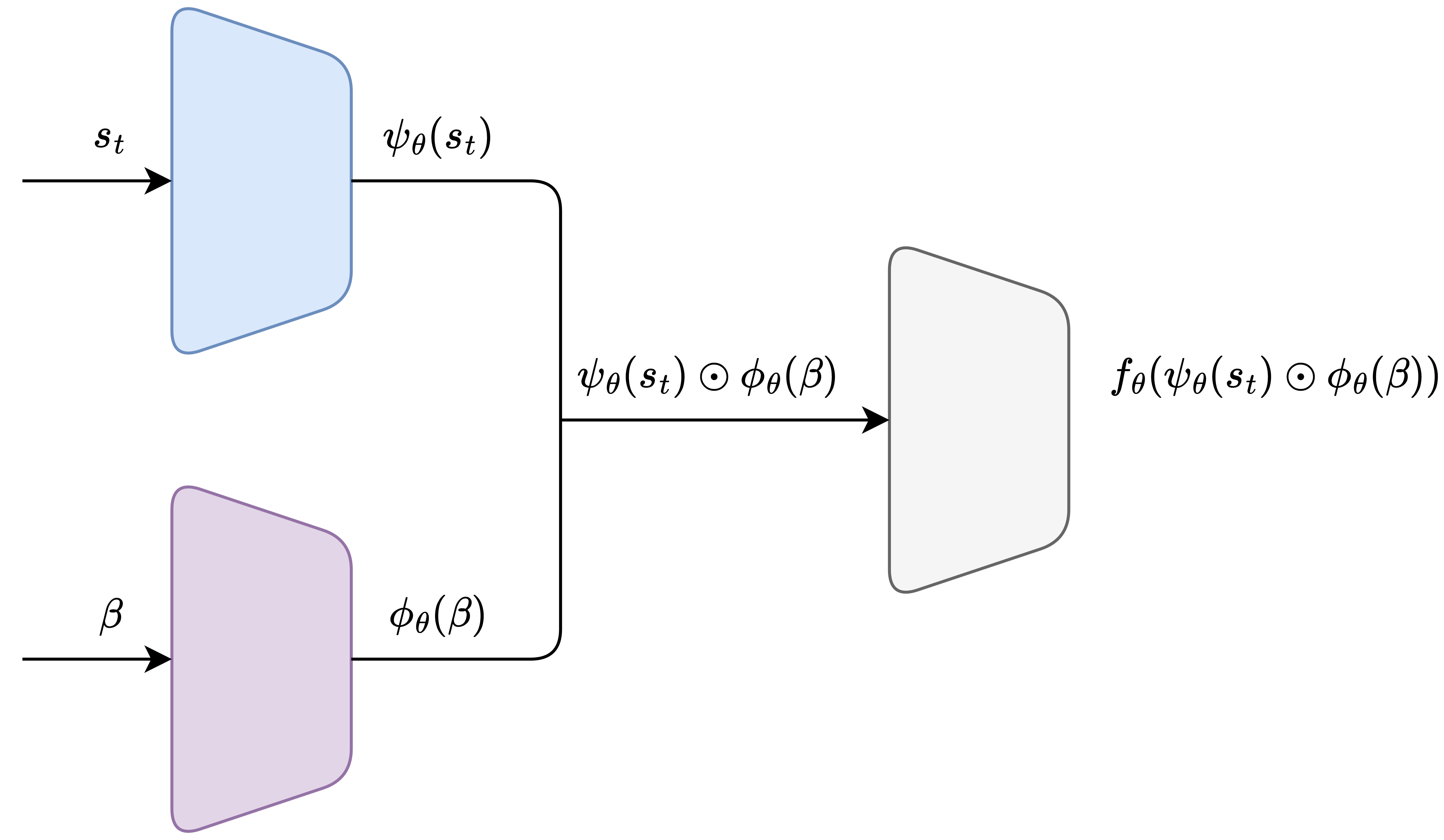}
	\caption{Illustration of the neural architecture used in IQN.}
	\label{iqn architecture}
\end{figure}

IQN \cite{dabney2018implicit} builds upon the ideas introduced in QR-DQN to estimate quantiles of the distribution associated with $Z(s_{t}, a_{t})$. However, in IQN the quantile levels $\beta$ are not fixed but are instead sampled from $ U([0, 1])$ at each training step. The notation $U([0, 1])$ denotes an uniform distribution over $[0, 1]$. Specifically, IQN uses a model $T_{\theta}$ such that $T_{\theta}(s_{t}, a_{t}, \beta)$ aims to estimate the $\beta$-quantile of the distribution associated to $Z(s_{t}, a_{t})$.

As illustrated in Figure \ref{iqn architecture}, a specific architecture is used to achieve this task. More precisely, we define $T_{\theta}(s_{t}, a_{t}, \beta) = f_{\theta} \left( \psi_{\theta}(s_{t}) \odot \phi_{\theta}(\beta) \right)_{a_{t}}$, where $\odot$ denote the element-wise (Hadamard) product. The model $\psi_{\theta}$ is composed of two stacked LSTM layers with $128$ and $64$ units, followed by a fully connected layer of $32$ units. Additionally, $\phi_{\theta}$ is used to embed the sampling $\beta$ using one dense fully connected layer applied to the vector $(cos(\pi \beta), cos(2\pi \beta),$ ... $
cos(n \pi \beta))$, where $n$ is a fixed number called the embedding dimension and set to $32$ in our experiments. Finally, $f_{\theta}$ consists of a fully connected layer using $32$ units. These models have been trained using a batch size of $16$.

The network $T_{\theta}$ is trained using the same loss function as in QR-DQN
\begin{align}
        \sum_{i=1}^{N} \mathbb{E}_{\beta_{j}} & \big[ 
    \rho_{\beta_{i}}^{k} \left(  r_{t} + \gamma T_{\overline{\theta}}(s_{t+1}, a^{\ast}, \beta_{j})
    - T_{\theta}(s_{t}, a_{t}, \beta_{i})
    \right)
    \big]  \nonumber \\
    & \simeq \frac{1}{N'} \sum_{i=1}^{N} \sum_{j=1}^{N'} \rho_{\beta_{i}}^{k} \left(  r_{t} + \gamma T_{\overline{\theta}}(s_{t+1}, a^{\ast}, \beta_{j})
    - T_{\theta}(s_{t}, a_{t}, \beta_{i})
    \right)
\end{align}
where $N$ and $N'$ denote the number of independent and identically distributed (iid) samples $\beta_{i}, \beta_{j} \sim U([0, 1])$. During our experiments, we used a value of $64$ for $N$ and $32$ for $N'$. The action $a^{\ast}$ is chosen according to the associated policy $a^{\ast} = \pi_{IQN}(s_{t+1})$ which is defined as
\begin{equation} \label{iqn policy}
    \pi_{IQN}(s_{t+1}) = \text{argmax}_{a  \in \mathcal{A}} \text{ } \frac{1}{K} \sum_{k=1}^{K} T_{\theta}(s_{t+1}, a, \beta_{k})
\end{equation}
whereas usually, the $\beta_{k}$ come from an iid sampling of $U([0, 1])$. In our experiments, we used a $K$ value of $64$. 
\medskip
\newline
For building risk-sensitive policies IQN$_{\alpha}$, that chooses actions to maximize CVaR, we only need to change the sampling distribution of the Equation \ref{iqn policy} to $U\left([0, \alpha] \right)$. The model $\psi_{\theta}$ used by IQN$_{\alpha}$ is composed of two stacked LSTM layers of $128$, and $64$ units respectively, followed by a dense layer of $32$.The model $f_{\theta}$ is a dense layer using $32$ units and the embedding dimension is also set to $32$. Finally, these models have been trained using a batch size of $32$ and a learning rate of $1\times 10^{-4}$.

\section{Experiments} \label{Experiments}

\subsection{Experimental setup}

In this study, we employ policies presented in the previous section to trade natural gas futures. These contracts are traded actively on a daily basis and are primarily distinguished by their maturity. Contracts with the nearest expiration are commonly referred to as front-month futures. For the sake of simplicity and due to an observed high correlation between front-month and other contracts, our agent will exclusively engage in trading front-month futures. To mitigate the risk of physical delivery and address potential market liquidity concerns, it is common to transition from the front-month contract to the next one, a fixed number of days before its expiration. In this study, we adhere to this practice and set this number to seven. Additionally, in our experiments, an episode consists of five consecutive trading days, and at the beginning of each episode, the agent does not hold any position.

A dataset provided by Predictive Layer SA that covers a period from the beginning of $2017$ to the end of $2022$ has been employed to build the MDP used to model agent-environment interactions. This dataset contains daily features related to futures traded at the Dutch Title Transfer Facility (TTF) market.

The TTF emerges as the leading European gas trading hub, characterized by its extensive range of available products and substantial trading volumes. In 2019, TTF trading accounted for an overwhelming 79\% of the total traded volumes in Europe \cite{heather2020european}. Notably, the TTF stands as the unique European hub exhibiting high liquidity across the medium- to long-term price curve. This pronounced liquidity attracts market participants throughout Europe to employ the TTF for arbitrage or hedging purposes, irrespective of whether the underlying asset originates from \cite{report2022}.

Now, we will delve into the specification of each component of the MDP used in our experiments.
\medskip
\newline
\textbf{Environment.} Each environment state $s_{t}$ is represented as a sequence $[\hat{s}_{t-9}, \hat{s}_{t-8}, \ldots \hat{s}_{t}]$, where $\hat{s}_{t}$ comprises two components denoted as $x_{t}$ and $c_{t}$. To construct $x_{t}$, we use the following procedure. Initially, we extract features $o_{t}$ related to the front-month futures from the dataset provided by Predictive Layer SA. In the cases where the dataset lacked features related to the front-month contract, we use the first available feature sorted according to the expiration dates of the associated futures. Each feature forms a vector of dimension $1141$ containing many statistics about the historical futures prices, such as the candlestick charts for instance.
Additionally, $o_{t}$ incorporates information related to other financial commodities, weather statistics associated with key European cities, and commodity fundamentals, such as gas storage levels in Europe. These time series have been stationarized using fractional differentiation \cite{walasek2021fractional} and further processed with common techniques like rank and smoothing methods, using sliding windows. Subsequently, this vector undergoes encoding via Principal Component Analysis (PCA) \cite{wold1987principal}. Several trials were conducted using DQN to determine the optimal dimension reduction. The dimensionality yielding the best results was found to be $75$, which was subsequently adopted across all agents. Finally, $\Delta_{t}$, the price difference of the front-month contract between business days $t$ and $t-1$ expressed in EUR/MWh, is incorporated to complete the construction of $x_{t}$.

The second component $c_{t}$ indicates the current position held by the agent. Specifically, a positive $c_{t}$ implies the agent is long for $c_{t}$ futures on day $t$. Conversely, a negative  $c_{t}$ indicates that the agent has a short position for $-c_{t}$ contracts. We constrain $c_{t}$ to fall within the range of $-10$ to $10$.

To adhere to the methodology employed by Predictive Layer SA to use the Extra-Trees model, we make slight adjustments to the inputs of this agent. Specifically, this model does not take a sequence as input, instead, it is fed with the complete feature $o_{t}$ without undergoing any reduction.
\medskip
\newline
\textbf{Actions.} Each action $a_{t}$ represents the number of contracts the agent buys or sells for a given business day, constrained by a maximum limit of three contracts. Agent's position for the next time step is updated as follows: $c_{t+1} = \text{min} \{ 10, \text{max}\{c_{t} + a_{t}, -10 \} \}$. Finally, remark that the Extra-Trees model does not output action $a_{t}$ as described above but binary prediction $\hat{y}_{t}$. These predictions are translated into actions using $a_{t} = 3\hat{y}_{t}$.
\medskip
\newline
\textbf{Rewards.} The considered reward function is derived from multiplying the current position held by the agent by the Sharpe Ratio \cite{sharpe1998sharpe} of the return $\Delta_{t+1}$. We assumed a risk-free return of zeros, and therefore, we calculate the Sharpe Ratio by dividing $\Delta_{t+1}$ with the standard deviation of the last observed price differences. In this paper, we use the ten most recent observed price differences. In other words, the reward function is defined as $R(s_{t}, a_{t}) = c_{t} \frac{\Delta_{t+1}}{\sigma_{t+1}}$, where $\sigma_{t+1}$ is the standard deviation of the returns $\Delta_{t-8}, \ldots , \Delta_{t+1}$.
\medskip
\newline
\textbf{Transition probability.}  The process used for transitioning between consecutive states is straightforward. We sequentially iterate through each business day to construct the state $s_{t+1}$ using $o_{t+1}$ and the position $c_{t+1}$.
\medskip
\newline
\textbf{Discount factor.} In our experiments, we fix the discounted factor $\gamma$ at $0.9$.
\medskip
\newline

\begin{figure}
	\centering
		\includegraphics[width = 0.9\linewidth]{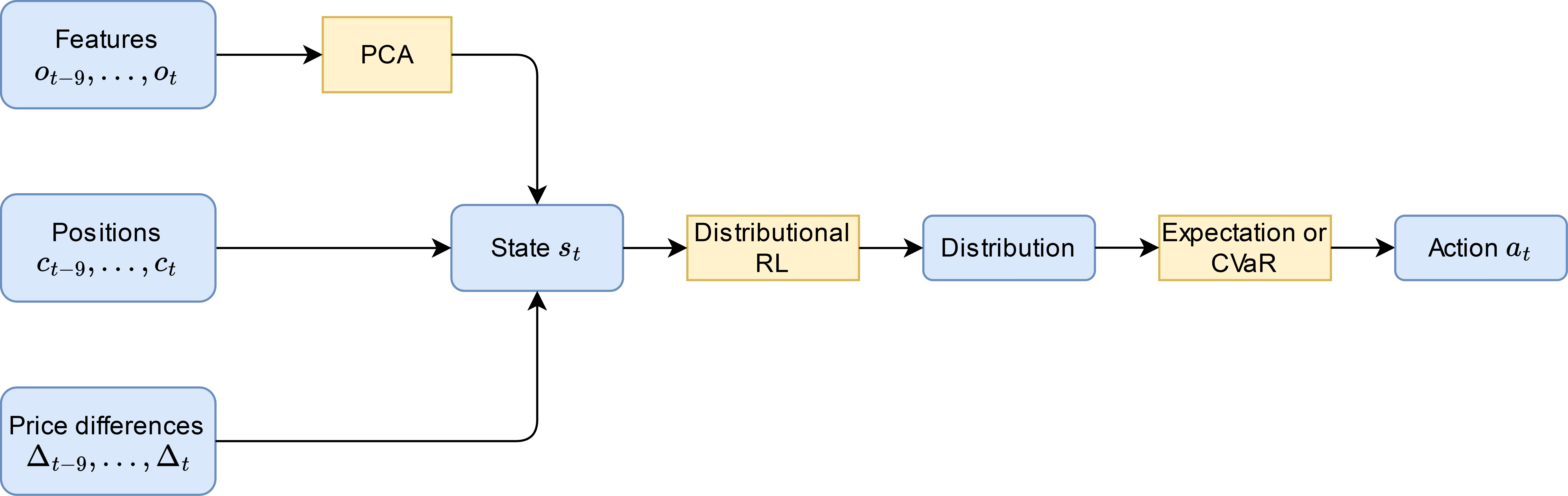}
	\caption{Overview of the proposed approach illustrating states constructions and their subsequent use.}
	\label{Pipeline}
\end{figure}

Figure \ref{Pipeline} provides an overview of the proposed approach, illustrating the construction of states from input features, positions, and price differences, along with their subsequent utilization.

\subsection{Evaluation process}

Our experiments are designed to study two distinct aspects of each policy: performance and risk aversion. In this section, we present the methodology employed to measure these characteristics.

\begin{figure}
	\centering
		\includegraphics[width = 12cm]{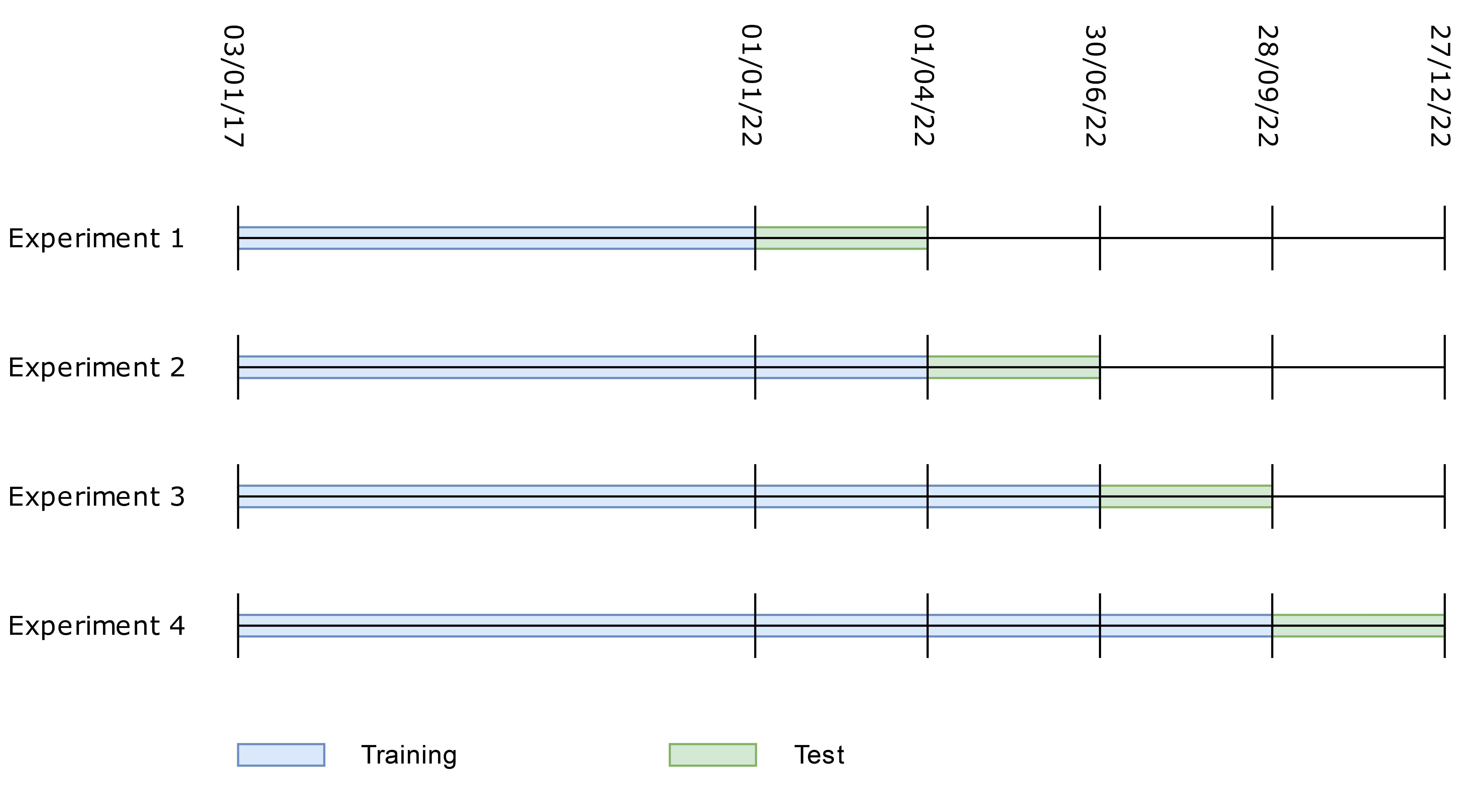}
	\caption{Illustration of training and testing periods employed in the four conducted experiments.}
	\label{cvexperiments}
\end{figure}

\subsubsection{Performance}

One experiment conducted to evaluate policy performance uses the following steps. Initially, a fixed period of the data is designated for training, while another period is reserved for testing. The test period spans 90 consecutive days, however, not all of these are business days, resulting in the actual testing period being shorter than 90 days. Test episodes are non-overlapping, ensuring that data related to a specific day within the test period is utilized only once. Finally, the cumulative Profit and Loss (P\&L) achieved across all test episodes during the specific period is measured. In other words, the performance of the model is computed using $\sum_{t} c_{t} \Delta_{t+1}$ where the sum is taken over all trading days of the test period. Recall that $\Delta_{t+1}$ represents the price difference between business days $t+1$ and $t$ of the front-month contracts. Thus, this metric accurately reflects the economic gain achieved by the agent, assuming zero transaction costs.

\begin{figure}
	\centering
		\includegraphics[width = 6cm]{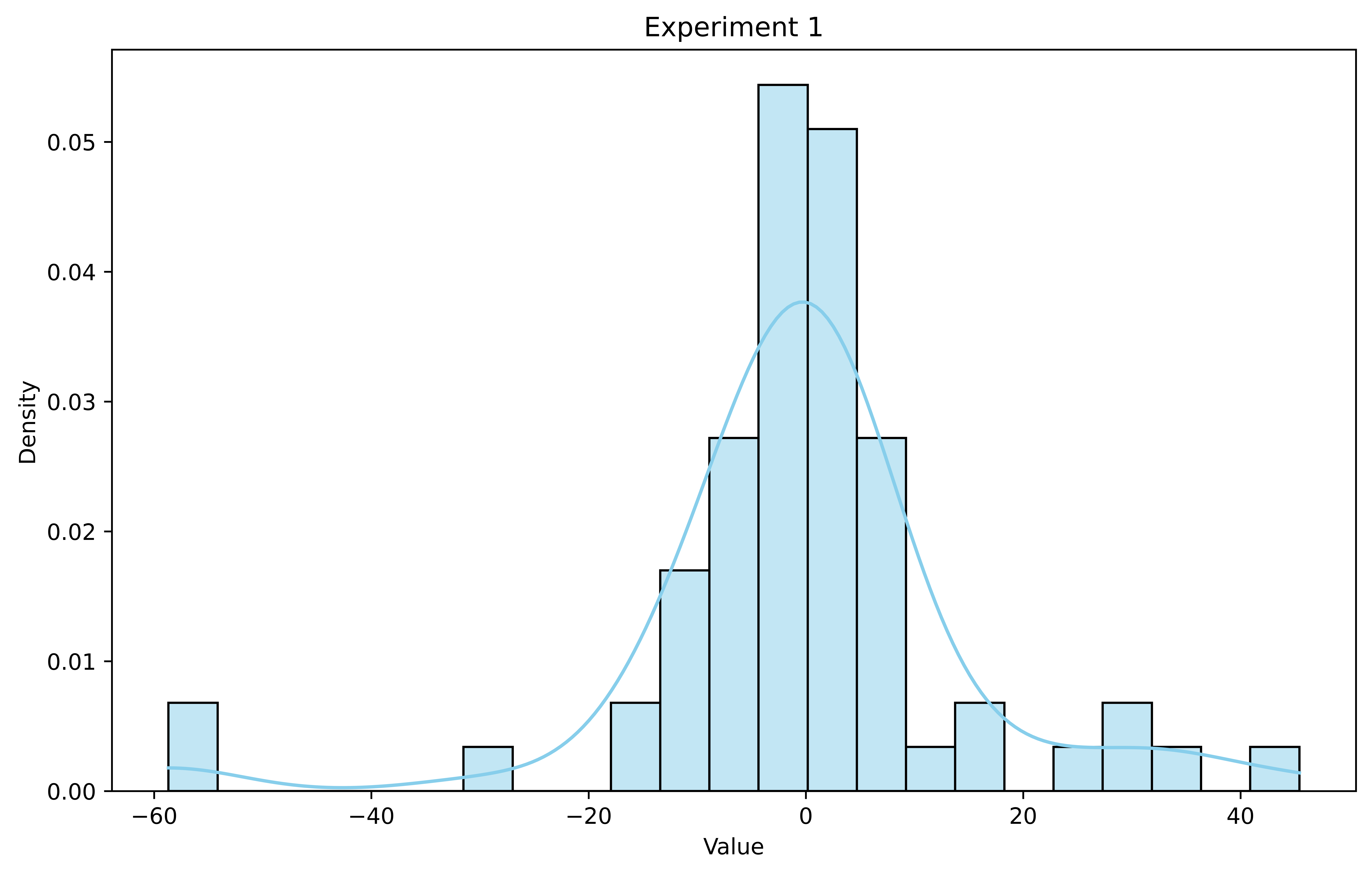}
        \includegraphics[width = 6cm]{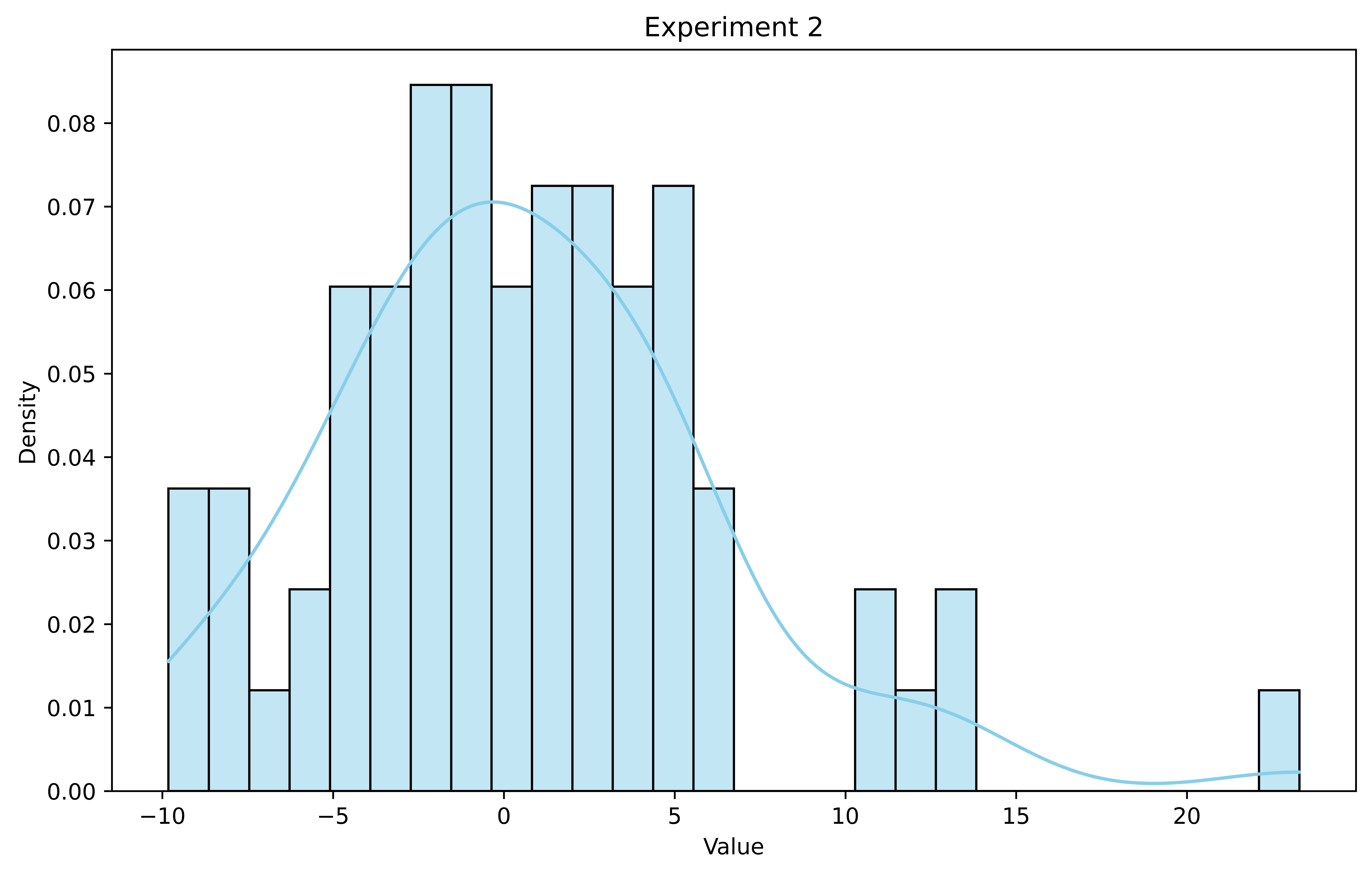}
        \includegraphics[width = 6cm]{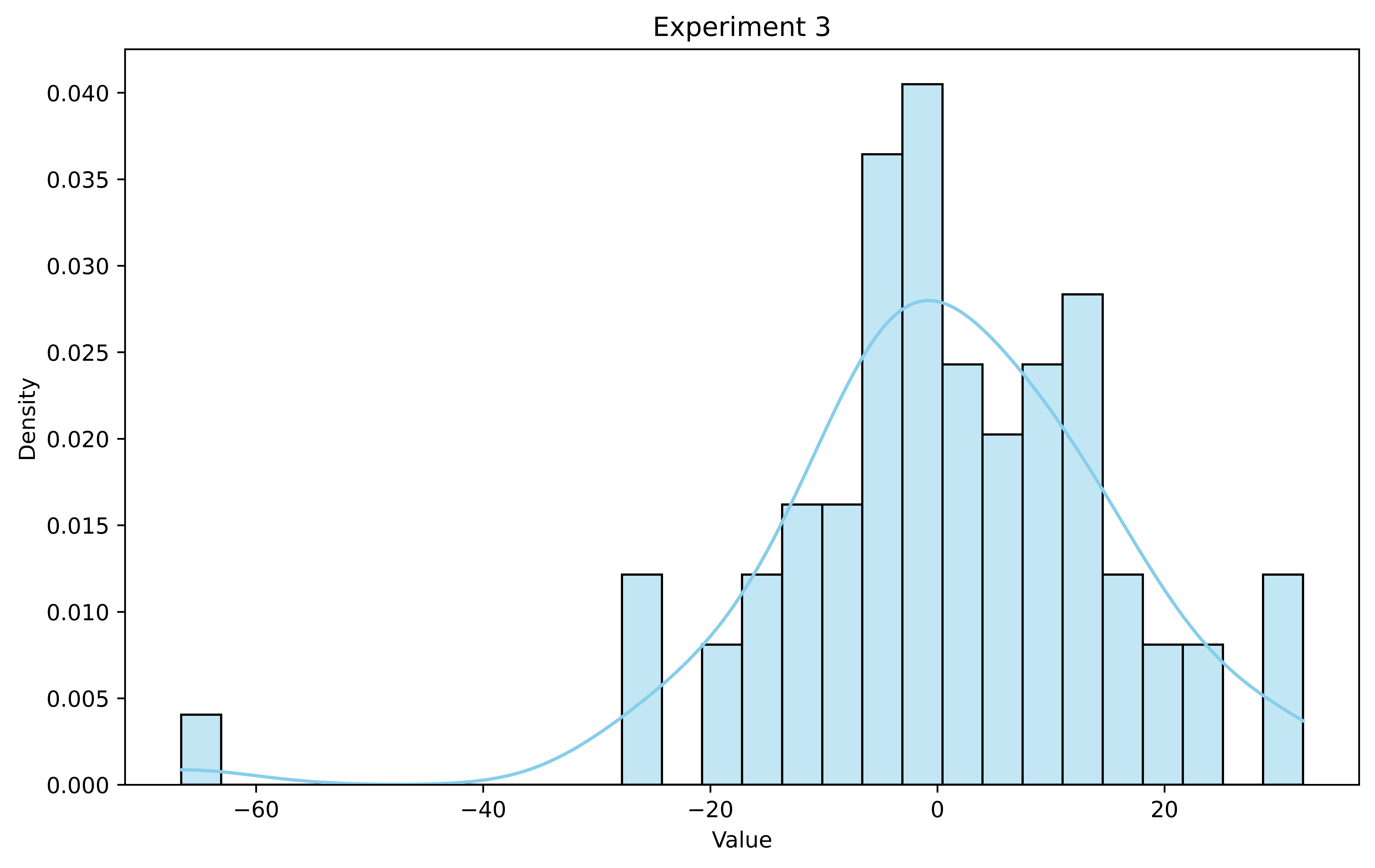}
        \includegraphics[width = 6cm]{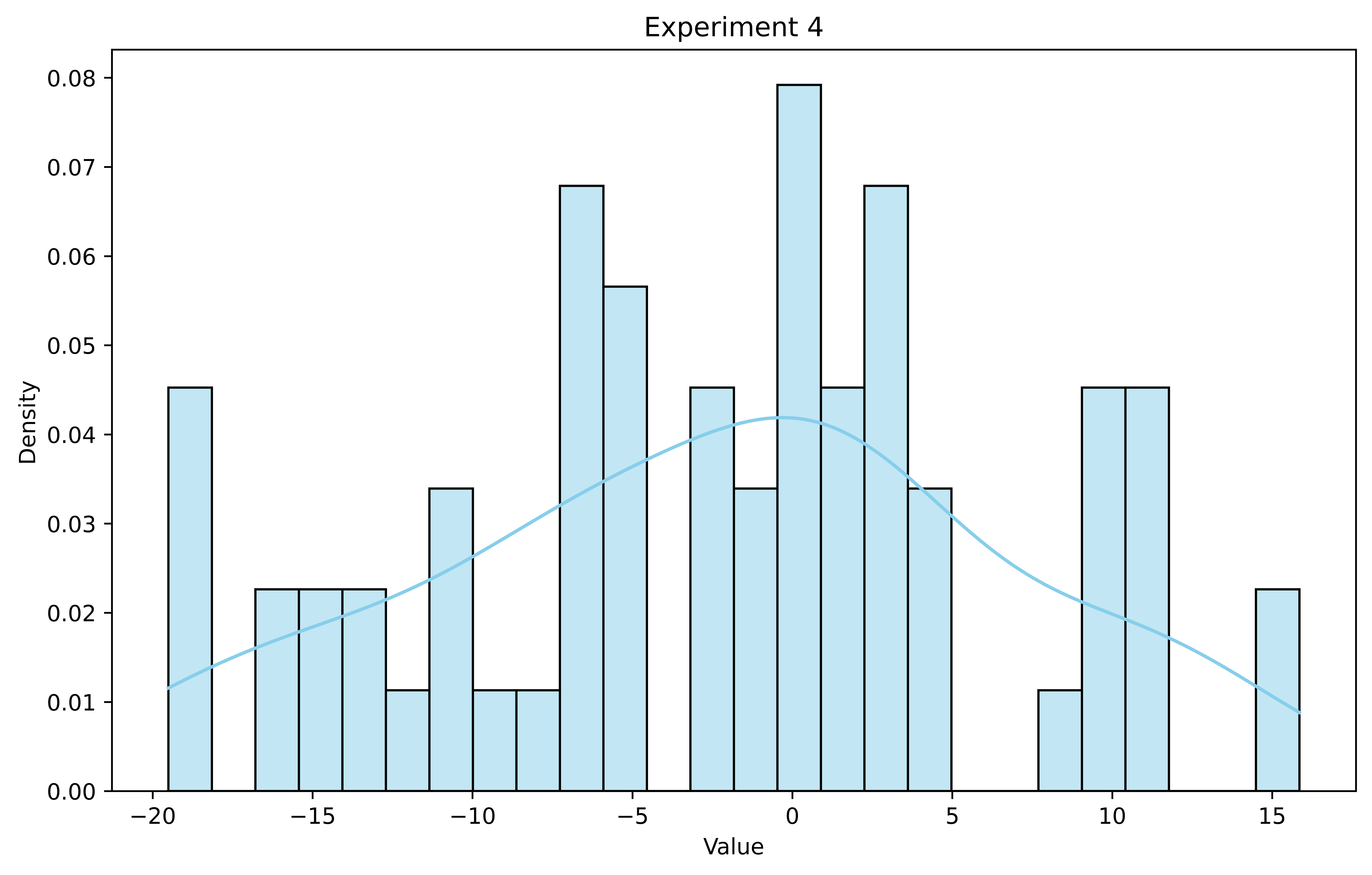}
	\caption{Distributions of $\Delta_{t}$ with corresponding kernel density estimates across the four testing conditions.}
	\label{Tests distribution}
\end{figure}

To ensure relevant evaluation, we conduct four experiments. As illustrated in Figure \ref{cvexperiments}, each experiment employs different training and testing periods. However, identical hyperparameters are utilized across all experiments. As shown in Figure \hypersetup{linkcolor=red}{\ref{Tests distribution}}, which presents the distributions of $\Delta_{t}$ along with corresponding kernel density estimate \cite{chen2017tutorial} across the four testing conditions, a significant distributional shift can be observed between the different experimented scenarios.

\subsubsection{Risk aversion}

In this study, our interest extends beyond policy performance to include their risk-averse behavior. To measure this characteristic, we incorporate the following process in each experiment. First, we define a \textit{risky} state as a state where the policy holds a long or short position for at least $7$ contracts and where the volatility $\sigma_{t}$ exceeds a predefined threshold $\hat{\sigma}$.
We assess policy risk aversion by quantifying the percentage of risky states encountered compared to the riskiest possible policies, characterized by consistently taking long or short positions with the maximum authorized number of contracts. For our experiments, we fix $\hat{\sigma}$ so that approximately $40\%$ of the states encountered by the riskiest policies are classified as risky.

\subsection{Results}\label{results}

First, Table \ref{Performance overview} provides a comprehensive summary of trading agent performances, encompassing both risk-free and risk-sensitive policies trained with a confidence level of $0.7$. Distributional RL algorithms exhibit superior results compared to non-distributional agents. Notably, C51 achieves an average P\&L that surpasses all non-distributional RL policies by more than $32\%$. Although the performance of C51 decreases with the adoption of its risk-sensitive variant, C51$_{0.7}$ still outperforms the risk-free policies. Furthermore, it can be observed that both IQN$_{0.7}$ and QR-DQN$_{0.7}$ outperform their respective risk-free counterparts.


\begin{table}[h]
\caption{P\&L of risk-free and risk-sensitive policies trained with a confidence level of $0.7$. We bold the highest mean and highest performance for each experiment.}\label{Performance overview}
\begin{tabular*}{\textwidth}{@{\extracolsep{\fill}} l l l l l l}
\hline
Agent & Exp. 1 & Exp. 2 & Exp. 3 & Exp. 4 & Mean \\
\hline
Extra-Trees & $65.37$ & $401.3$ & $758.5$ & $268.7$ & $373.47$ \\
DQN & $460.3$ & $266.7$ & $899.4$ & $1028$ & $663.6$  \\
Prioritized DQN & $304.4$ & $190.6$ & $634.4$ & $\textbf{1493}$ & $655.6$ \\
Dueling DQN & $1274$ & $361$ & $622.2$ & $-4.27$ & $563.23$ \\
Prioritized Dueling DQN & $1292$ & $288.3$ & $144.9$ & $655.7$ & $595.22$ \\
C51 & $1287$ & $51.35$ & $762.1$ & $1427$ & $\textbf{881.86}$ \\
C51$_{0.7}$ & $1382$ & $78.02$ & $\textbf{1367}$ & $279.4$ & $776.6$ \\
QR-DQN &  $1612$ & $239.7$ & $764.5$ & $108.1$ & $681.08 $ \\
QR-DQN$_{0.7}$ & $\textbf{1659}$ & $-9.201$ & $984$ & $860$ & $873.45 $ \\
IQN & $1308$ & $\textbf{567.5}$ & $464$ & $325.3$ & $666.2$ \\
IQN$_{0.7}$ & $1343$ & $258.4$ & $206.9$ & $927.4$ & $683.92$ \\
\hline
\end{tabular*}
\end{table}

\begin{table}[h]
\caption{Percentage of risky states encountered by risk-free and risk-sensitive policies trained with a confidence level of $0.7$. We bold the lowest mean and lowest percentage for each experiment.}\label{risk-averse overview}
\begin{tabular*}{\textwidth}{@{\extracolsep{\fill}} l l l l l l}
\hline
Agent & Exp. 1 & Exp. 2 & Exp. 3 & Exp. 4 & Mean \\
\hline
Extra-Trees & $46.43$ & $26.67$ & $50.0$ & $35.48$ & $39.64$ \\
DQN & $\textbf{10.71}$ & $46.67$ & $50.0$ & $58.06$ & $41.36$ \\
Prioritized DQN & $\textbf{10.71}$ & $33.33$ & $30.0$ & $35.48$ & $\textbf{27.38}$ \\
Dueling DQN & $57.14$ & $40.0$ & $82.5$ & $48.39$ & $57.01$ \\
Prioritized Dueling DQN & $57.14$ & $13.33$ & $100.0$ & $16.13$ & $46.65$ \\
C51 & $25.0$ & $\textbf{0.0}$ & $85.0$ & $61.29$ & $42.82$ \\
C51$_{0.7}$ & $39.29$ & $53.33$ & $27.5$ & $\textbf{6.45}$ & $31.64$ \\
QR-DQN &  $57.14$ & $86.67$ & $\textbf{5.0}$ & $51.61$ & $50.1 $ \\
QR-DQN$_{0.7}$ & $64.29$ & $66.67$ & $52.5$ & $83.87$ & $66.83 $ \\
IQN & $57.14$ & $53.33$ & $77.5$ & $54.84$ & $60.7$ \\
IQN$_{0.7}$ & $60.71$ & $66.67$ & $10.0$ & $41.94$ & $44.83$ \\
\hline
\end{tabular*}
\end{table}
\noindent
Moreover, an examination of the risk-averse tendencies exhibited by both risk-free and risk-sensitive policies trained with a confidence level of $0.7$ is detailed in Table \ref{risk-averse overview}. Risk aversion exhibits significant variability across the considered agents, ranging from a mean of encountered risky states of $27.38\%$ with Prioritized DQN to $66.83\%$ with QR-DQN$_{0.7}$. Notably, QR-DQN$_{0.7}$ and IQN display the lowest levels of risk aversion, with mean scores of $66.83\%$ and $60.7\%$, respectively. Unexpectedly, it can be noted that QR-DQN presents a higher risk aversion than QR-DQN$_{0.7}$. In contrast, adopting the risk-sensitive variants of C51 and IQN enables them to increase their risk aversion. Specifically,  IQN$_{0.7}$ enables a reduction of $15.87\%$, and C51 of $11.18\%$.
\begin{table}[h]
\caption{Percentage of risky states encountered by C51$_{\alpha}$. We bold the lowest mean and lowest percentage for each experiment.}\label{C51 risk}
\begin{tabular*}{\textwidth}{@{\extracolsep{\fill}} l l l l l l}
\hline
Agent & Exp. 1 & Exp. 2 & Exp. 3 & Exp. 4 & Mean \\
\hline
C51 & $25.0$ & $\textbf{0.0}$ & $85.0$ & $61.29$ & $42.82$ \\
C51$_{0.9}$ & $60.71$ & $53.33$ & $30.0$ & $\textbf{0.0}$ & $36.01$ \\
C51$_{0.7}$ & $39.29$ & $53.33$ & $\textbf{27.5}$ & $6.45$ & $31.64$  \\
C51$_{0.5}$ & $\textbf{0.0}$ & $26.67$ & $62.5$ & $\textbf{0.0}$ & $22.29$  \\
C51$_{0.3}$ & $50.0$ & $\textbf{0.0}$ & $40.0$ & $48.39$ & $34.6$  \\
C51$_{0.1}$ & $\textbf{0.0}$ & $\textbf{0.0}$ & $52.5$ & $3.23$ & $\textbf{13.93}$  \\
\hline
\end{tabular*}
\end{table}

\begin{table}[h]
\caption{ P\&L of C51$_{\alpha}$. We bold the highest mean and highest performance for each experiment.}\label{C51 performance}
\begin{tabular*}{\textwidth}{@{\extracolsep{\fill}} l l l l l l}
\hline
Agent & Exp. 1 & Exp. 2 & Exp. 3 & Exp. 4 & Mean \\
\hline
C51 & $1287$ & $51.35$ & $762.1$ & $\textbf{1427}$ & $\textbf{881.86}$  \\
C51$_{0.9}$ & $1439$ & $\textbf{125.4}$ & $397.2$ & $420.4$ & $595.5$ \\ 
C51$_{0.7}$ & $\textbf{1382}$ & $78.02$ & $\textbf{1367}$ & $279.4$ & $776.6$ \\
C51$_{0.5}$ & $330$ & $43.82$ & $1255$ & $-36.76$ & $398.02$ \\
C51$_{0.3}$ & $639.3$ & $-98.66$ & $-142.7$ & $568.2$ & $241.54$ \\
C51$_{0.1}$ & $286.9$ & $0$ & $755.4$ & $-469.8$ & $143.12$  \\
\hline
\end{tabular*}
\end{table}
\medskip
\noindent
Furthermore, ablation studies have been conducted to investigate the influence of the confidence level on risk-sensitive policies. Interestingly, these results reveal variability depending on the algorithm under examination. First, as shown in Table \ref{C51 risk}, reducing the confidence level generally enhances the risk aversion of C51$_{\alpha}$.  However, this trend exhibits an exception, observed with C51$_{0.3}$, which deviates from this pattern. Suprisingly, C51$_{0.3}$ exhibits a risk-averse behavior comparable to C51$_{0.9}$. Furthermore, Table \ref{C51 performance} indicates a correlation between the policy risk aversion and its mean P\&L. Lower risk aversion tends to yield higher performance. Consequently, it is unsurprising that the worst result is achieved by C51$_{0.1}$, the policy with the highest risk aversion, while the best results are obtained by C51, characterized by the lowest level of risk aversion.
\begin{table}[h]
\caption{Percentage of risky states encountered by QR-DQN$_{\alpha}$. We bold the lowest mean and lowest percentage for each experiment.}\label{QR-DQN risk}
\begin{tabular*}{\textwidth}{@{\extracolsep{\fill}} l l l l l l}
\hline
Agent & Exp. 1 & Exp. 2 & Exp. 3 & Exp. 4 & Mean \\
\hline
QR-DQN & $57.14$ & $86.67$ & $\textbf{5.0}$ & $\textbf{51.61}$ & $\textbf{50.1}$ \\
QR-DQN$_{0.9}$ & $53.57$ & $60.0$ & $57.5$ & $58.06$ & $57.28 $ \\
QR-DQN$_{0.7}$ &  $64.29$ & $66.67$ & $52.5$ & $83.87$ & $66.83 $ \\
QR-DQN$_{0.5}$ & $71.43$ & $66.67$ & $55.0$ & $74.19$ & $66.82 $ \\
QR-DQN$_{0.3}$ & $67.86$ & $\textbf{40.0}$ & $50.0$ & $83.87$ & $60.43 $ \\
QR-DQN$_{0.1}$ & $\textbf{42.86}$ & $60.0$ & $60.0$ & $93.55$ & $64.1 $ \\
\hline
\end{tabular*}
\end{table}

\begin{table}[h]
\caption{P\&L of QR-DQN$_{\alpha}$. We bold the highest mean and highest performance for each experiment.}\label{QR-DQN performance}
\begin{tabular*}{\textwidth}{@{\extracolsep{\fill}} l l l l l l}
\hline
Agent & Exp. 1 & Exp. 2 & Exp. 3 & Exp. 4 & Mean \\
\hline
QR-DQN &  $1612$ & $239.7$ & $764.5$ & $108.1$ & $681.08 $ \\
QR-DQN$_{0.9}$ & $1345$ & $246.8$ & $1303$ & $112.4$ & $751.8 $ \\
QR-DQN$_{0.7}$ & $\textbf{1659}$ & $-9.201$ & $984$ & $860$ & $873.45 $ \\
QR-DQN$_{0.5}$ & $1351$ & $\textbf{371.3}$ & $510.9$ & $517.5$ & $687.68 $ \\
QR-DQN$_{0.3}$ & $1312$ & $53.59$ & $\textbf{1515}$ & $855.1$ & $\textbf{933.92}$ \\
QR-DQN$_{0.1}$ & $1289$ & $177.6$ & $223.2$ & $\textbf{995.9}$ & $671.42 $ \\
\hline
\end{tabular*}
\end{table}
\noindent
As presented in Table \ref{QR-DQN risk}, the results concerning QR-DQN$_{\alpha}$ diverge from those observed with C51$_{\alpha}$. Specifically, QR-DQN$_{\alpha}$ does not exhibit a clear relationship between reductions in the confidence level and changes in risk-averse behavior. Unexpectedly, QR-DQN demonstrates the highest levels of risk aversion, while QR-DQN$_{0.7}$ displays the lowest. Additionally, QR-DQN$_{0.7}$, QR-DQN$_{0.5}$, QR-DQN$_{0.1}$ exhibit comparable risk-averse behavior. As shown in Table \ref{QR-DQN performance}, the performances of these risk-sensitive policies also indicate that decreasing the confidence level does not necessarily correspond to a reduction in average P\&L. It can be observed that QR-DQN$_{0.3}$, QR-DQN$_{0.7}$ and QR-DQN$_{0.9}$ deliver the best three performances, while QR-DQN$_{0.1}$ and QR-DQN$_{0.5}$ present results comparable with QR-DQN.
\begin{table}[h]
\caption{Percentage of risky states encountered by IQN$_{\alpha}$. We bold the lowest mean and lowest percentage for each experiment.}\label{IQN risk}
\begin{tabular*}{\textwidth}{@{\extracolsep{\fill}} l l l l l l}
\hline
Agent & Exp. 1 & Exp. 2 & Exp. 3 & Exp. 4 & Mean \\
\hline
IQN & $57.14$ & $53.33$ & $77.5$ & $54.84$ & $60.7$ \\
IQN$_{0.9}$ & $57.14$ & $80.0$ & $47.5$ & $48.39$ & $58.26$ \\
IQN$_{0.7}$ & $60.71$ & $66.67$ & $10.0$ & $41.94$ & $44.83$ \\
IQN$_{0.5}$ & $60.71$ & $66.67$ & $0.0$ & $0.0$ & $31.84$ \\
IQN$_{0.3}$ & $\textbf{46.43}$ & $\textbf{0.0}$ & $5.0$ & $\textbf{0.0}$ & $\textbf{12.86}$ \\
IQN$_{0.1}$ & $85.71$ & $\textbf{0.0}$ & $\textbf{0.0}$ & $\textbf{0.0}$ & $21.43$ \\
\hline
\end{tabular*}
\end{table}

\begin{table}[h]
\caption{P\&L of IQN$_{\alpha}$. We bold the highest mean and highest performance for each experiment.}\label{IQN performance}
\begin{tabular*}{\textwidth}{@{\extracolsep{\fill}} l l l l l l}
\hline
Agent & Exp. 1 & Exp. 2 & Exp. 3 & Exp. 4 & Mean \\
\hline
IQN & $1308$ & $\textbf{567.5}$ & $464$ & $325.3$ & $666.2$  \\
IQN$_{0.9}$ & $\textbf{1461}$ & $199$ & $\textbf{989.6}$ & $378.7$ & $\textbf{757.08}$ \\
IQN$_{0.7}$ & $1343$ & $258.4$ & $206.9$ & $\textbf{927.4}$ & $683.92$  \\
IQN$_{0.5}$ & $1143$ & $285.3$ & $0$ & $0$ & $357.08$ \\
IQN$_{0.3}$ & $636.3$ & $-141$ & $569.5$ & $0$ & $266.2$ \\
IQN$_{0.1}$ & $836.9$ & $0$ & $0$ & $0$ & $209.22$ \\
\hline
\end{tabular*}
\end{table}
In contrast to QR-DQN$_{\alpha}$, and as evidenced in Table \ref{IQN risk}, reducing confidence level consistently enhances the risk aversion of IQN$_{\alpha}$. The only exception is observed with IQN$_{0.1}$, which displays lower risk aversion than IQN$_{0.3}$. However, it is noteworthy that IQN$_{0.1}$ still demonstrates a high level of risk aversion underlined by a median of encountered risky states reaching $0.0$. Moreover, IQN$_{\alpha}$ showcases a broad spectrum of risk-averse behavior, ranging from $12.86\%$, the lowest score encountered in our experiments to $60.7\%$ one of the highest. Furthermore, as presented in Table \ref{IQN performance}, the performance of IQN$_{\alpha}$ aligns closely with its corresponding risk-averse behavior. Agents with higher risk aversion tend to exhibit lower P\&L. Finally, it is worth noting that the best result is achieved by IQN$_{0.9}$.

\section{Discussion} \label{Discussion}

\subsection{Challenges in policy development}
The results presented in the previous section unveil several insights. First, they underscore the intricate nature of constructing policies capable of delivering compelling performance across different settings. Indeed, nearly all policies yield unsatisfactory performance in at least one experiment. 
This challenge is further illustrated by the accuracy of the Extra-Trees model, which achieves a score of 44.62\%, 48.57\%, 52.86\%, and 38.46\% in Experiments 1, 2, 3, and 4, respectively. Despite these moderate accuracy levels, the Extra-Trees model produces relatively interesting cumulative P\&L, underlying that, in this context, accuracy may not be the most relevant performance metric. In particular, accurately predicting large market movements has a more substantial impact on economic outcomes than correctly forecasting smaller fluctuations. This underscores the need for policies that prioritize impactful predictions over general accuracy.

Finally, the complexity of constructing policies able to deliver compelling performance across different settings is also highlighted in Table \ref{Performance overview} by the observation that the top-performing policy varies across each experiment. However, it is worth noting that three out of these four models are distributional policies.

\subsection{Risk-aversion across risk-sensitive policies}
The results suggest that, unlike QR-DQN$_{\alpha}$, C51$_{\alpha}$ and IQN$_{\alpha}$ could potentially be employed to develop trading strategies tailored to specific levels of risk aversion. Notably, IQN$_{\alpha}$ appears particularly well-suited for this purpose, covering a wider range of risk-averse behaviors compared to C51$_{\alpha}$. Specifically, while the proportion of risky states encountered by IQN$_{\alpha}$ spans from $12.86\%$ to $60.7\%$, C51$_{\alpha}$ ranges from $13.93\%$ to $42.82\%$. Additionally, in situations where decreasing the confidence level does not enhance risk aversion, divergent behaviors can be observed between IQN$_{\alpha}$ and C51$_{\alpha}$. While IQN$_{\alpha}$ demonstrates a risk-averse behavior close to the expected level, as illustrated with IQN$_{0.1}$, C51$_{\alpha}$ exhibits a more hazardous behavior. For instance, C51$_{0.3}$ displays a risk aversion comparable to C51$_{0.9}$. These observations show that IQN$_{\alpha}$ cover a broader spectrum of risk-averse behavior while achieving more robust outcomes than C51$_{\alpha}$, indicating that IQN$_{\alpha}$ might be better suited for constructing risk-averse policies.

Furthermore, in contrast to C51, both IQN, and QR-DQN demonstrate an enhanced performance when training policies to maximize CVaR. We hypothesize that this performance improvement can be attributed to the dual representation of CVaR, which may enhance robustness across different training and testing conditions. Moreover, the utilization of quantile regression in IQN and QR-DQN may render them better suited for this task, thereby elucidating why C51 does not exhibit similar behavior.

\subsection{Role of Sharpe Ratio and normalization}
Now, recall that the reward function utilized in this study is based on the Sharpe Ratio. Since, this metric measures the return by a unit of risk, it might be tempting to assume that this choice is sufficient to build risk-averse policies. However, our results suggest otherwise. In fact, RL policies generally do not exhibit a higher level of risk aversion compared to Extra-Trees, which does not consider market volatility during the training process. Although the incorporation of the Sharpe Ratio into our reward function did not enhance risk aversion, it led to a notable improvement in performance. We hypothesize that these observations may be attributed to the Sharpe Ratio's role as a normalization method specifically tailored to our problem domain. This hypothesis finds support in several observations.

First, observe that the Sharpe Ratio corresponds to the z-score \cite{al2006normalization} when employing a sliding window to compute the mean and standard deviation. Additionally, the integration of a sliding window into the z-score normalization method might enable better handling of potential non-stationarity, thereby enhancing performance over conventional normalization methods in some specific situations \cite{tanaka2022sliding}. Furthermore, our initial attempts to train DQN policies using a reward function based on the P\&L yielded poor performances. Even considering the z-score of the P\&L did not show significant improvements. Our first promising results emerged only after integrating the Sharpe Ratio into the reward function, suggesting its pivotal role in achieving noteworthy performances.

In summary, the significant enhancement in policy performance resulting from the integration of the Sharpe Ratio into our reward function without exhibiting a risk-averse behavior along with the close relationship between the Sharpe Ratio and the z-score, prompts us to consider the Sharpe Ratio as a relevant normalization method tailored specifically to our problem.

\subsection{Limitations and future works}

While the experimental results highlight the ability of distributional RL algorithms to construct efficient and risk-averse trading strategies, this study is not without limitations, which offers exciting future work opportunities.

First, the agents studied in this work have only been tested for trading natural gas futures contracts, and their behaviors in other financial markets remain to be explored.

Additionally, to address the high dimensionality of features associated with each front-month future, we employed PCA to obtain a more compact representation. However, prior studies have proposed techniques specifically designed for encoding high-dimensional states in RL \cite{gelada2019deepmdp, lee2020stochastic, zhangrep2021}. Exploring the use of these specialized methods as alternatives to PCA may yield further improvements.

Furthermore, the choice of CVaR as the risk measure, may not be aligned with all desired trading strategies. Future research could explore the behavior of policies trained with alternative risk measures, such as the Wang risk measure \cite{wang2000class}, to enhance adaptability and broaden the applicability of distributional RL for trading.

Finally, and as noted earlier, most agents exhibit unsatisfactory performance in at least one experiment. An intriguing direction for future research could involve developing an ensemble learning approach \cite{zhou2012ensemble} that combines multiple RL policies to construct an algorithm capable of consistently delivering high performance across diverse settings. Another promising avenue could be integrating human expert knowledge into the method by incorporating an imitation learning component, potentially enhancing its robustness to different testing conditions.

\vspace{0.6cm}

\section{Conclusion} \label{Conclusion}

This study investigates the effectiveness of C51, QR-DQN, and IQN for
natural gas futures trading and explores their potential for constructing risk-averse policies. Four experiments were conducted using preprocessed data supplied by Predictive Layer SA. These policies were evaluated against five baselines, comprising four classical RL policies and one Extra-Trees model. The use of a meticulously constructed and preprocessed dataset sourced from a company active in energy trading, coupled with the comparison against a baseline provided by the same company, provides a relevant evaluation of policies' performance and behavior.

Our results suggest that distributional RL outperforms traditional RL algorithms, demonstrating a notable performance improvement of over $32\%$ with C51. Furthermore, our experiments show that distributional RL enables the development of risk-averse trading strategies by training policies to maximize CVaR instead of the expectation of the cumulative return. Finally, in contrast to QR-DQN$_{\alpha}$, using IQN$_{\alpha}$ or C51$_{\alpha}$, it becomes possible to create strategies that align with desired levels of risk aversion. However, IQN$_{\alpha}$ covers a wider spectrum of risk-averse behavior while achieving more robust outcomes than C51$_{\alpha}$, indicating that IQN$_{\alpha}$ might be better suited for constructing risk-averse policies. Furthermore, our results indicate that training policies to maximize CVaR, as opposed to the expected value, can potentially enhance the performance of both IQN and QR-DQN algorithms.

This study may attract the attention of financial professionals for several reasons. First, individuals looking for risk-averse portfolio management tend to avoid volatile markets, favoring low-volatility assets like bonds. However, this paper offers a novel approach to constructing trading strategies that mitigate exposition to high-risk positions in highly volatile markets. Therefore, this study presents new opportunities for improving the diversification of risk-averse portfolios. Moreover, our research highlights the potential of distributional RL algorithms to deliver impressive performances even in volatile market conditions.

While the approach introduced in this paper has been specifically tailored for natural gas futures trading, the underlying principles of using distributional RL to optimize risk measures like CVaR are not inherently restricted to finance. Furthermore, the results presented in this study underline the promise of this framework for other applications where managing uncertainty and developing risk-averse strategies are critical, such as supply chain management or energy systems.


\subsection*{Declaration of competing interest}
The authors declare that they have no known competing financial interests or personal relationships that could have appeared to influence the work reported in this paper.

\subsection*{Acknowledgments}
This work was supported by the InnoSuisse Project 54228.1 IP-ICT.

\begin{thebibliography}{100}

\bibitem{ali2020coronavirus}
Mohsin Ali, Nafis Alam, and Syed Aun~R Rizvi.
\newblock Coronavirus (covid-19)—an epidemic or pandemic for financial markets.
\newblock {\em Journal of Behavioral and Experimental Finance}, 27:100341, 2020.

\bibitem{zhang2020financial}
Dayong Zhang, Min Hu, and Qiang Ji.
\newblock Financial markets under the global pandemic of covid-19.
\newblock {\em Finance research letters}, 36:101528, 2020.

\bibitem{aruga2020effects}
Kentaka Aruga, Md~Monirul Islam, and Arifa Jannat.
\newblock Effects of covid-19 on indian energy consumption.
\newblock {\em Sustainability}, 12(14):5616, 2020.

\bibitem{jiang2021impacts}
Peng Jiang, Yee Van~Fan, and Ji{\v{r}}{\'\i}~Jarom{\'\i}r Kleme{\v{s}}.
\newblock Impacts of covid-19 on energy demand and consumption: Challenges, lessons and emerging opportunities.
\newblock {\em Applied energy}, 285:116441, 2021.

\bibitem{huang2023time}
Jionghao Huang, Baifan Chen, Yushi Xu, and Xiaohua Xia.
\newblock Time-frequency volatility transmission among energy commodities and financial markets during the covid-19 pandemic: A novel tvp-var frequency connectedness approach.
\newblock {\em Finance Research Letters}, 53:103634, 2023.

\bibitem{fang2022russia}
Yi~Fang and Zhiquan Shao.
\newblock The russia-ukraine conflict and volatility risk of commodity markets.
\newblock {\em Finance Research Letters}, 50:103264, 2022.

\bibitem{lin2024impact}
Yongjia Lin, Yizhi Wang, et~al.
\newblock The impact of the russia--ukraine war on volatility spillovers.
\newblock {\em International Review of Financial Analysis}, 93:103194, 2024.

\bibitem{ngwakwe2022stock}
Collins~C Ngwakwe.
\newblock Stock market volatility during rumours of war and actual war: Case of russia-ukraine conflict.
\newblock {\em Acta Universitatis Danubius. {\OE}conomica}, 18(3):55--70, 2022.

\bibitem{rice2023russia}
Brendan Rice, Manuel~A Hernandez, Joseph~W Glauber, and Rob Vos.
\newblock The russia-ukraine war is exacerbating international food price volatility.
\newblock {\em International Food Policy Research Institute}, 2023.

\bibitem{wu2023stock}
Feng-lin Wu, Xu-dong Zhan, Jia-qi Zhou, and Ming-hui Wang.
\newblock Stock market volatility and russia--ukraine conflict.
\newblock {\em Finance Research Letters}, 55:103919, 2023.

\bibitem{lo2022russo}
Gaye-Del Lo, Isaac Marcelin, Th{\'e}ophile Bass{\`e}ne, and Babacar S{\`e}ne.
\newblock The russo-ukrainian war and financial markets: the role of dependence on russian commodities.
\newblock {\em Finance Research Letters}, 50:103194, 2022.

\bibitem{virag2023turbulent}
Attila Vir{\'a}g and Gr{\'e}ta Tancsa.
\newblock Turbulent energy transformations in central europe: Nord stream projects in the context of geopolitics.
\newblock {\em Politics in Central Europe}, 19(1):113--144, 2023.

\bibitem{zhang2023tensions}
Yuanyuan Zhang.
\newblock Tensions between international economic law and the eu energy security regulations during the securitization of eu-russian gas relations: Way forward?
\newblock {\em Journal of World Trade}, 57(3), 2023.

\bibitem{report2022}
Jostein Kristensen.
\newblock The european gas market.
\newblock Technical report, Oxera, December 2022.
\newblock Report prepared for ICE.

\bibitem{enescu2023discussing}
Adrian-Gabriel Enescu and Monica~R{\u{a}}ileanu Szeles.
\newblock Discussing energy volatility and policy in the aftermath of the russia--ukraine conflict.
\newblock {\em Frontiers in Environmental Science}, 11:1225753, 2023.

\bibitem{inacio2023assessing}
CMC Inacio~Jr, L~Kristoufek, and SA~David.
\newblock Assessing the impact of the russia--ukraine war on energy prices: A dynamic cross-correlation analysis.
\newblock {\em Physica A: Statistical Mechanics and its Applications}, 626:129084, 2023.

\bibitem{brown2020language}
Tom Brown, Benjamin Mann, Nick Ryder, Melanie Subbiah, Jared~D Kaplan, Prafulla Dhariwal, Arvind Neelakantan, Pranav Shyam, Girish Sastry, Amanda Askell, et~al.
\newblock Language models are few-shot learners.
\newblock {\em Advances in neural information processing systems}, 33:1877--1901, 2020.

\bibitem{he2016deep}
Kaiming He, Xiangyu Zhang, Shaoqing Ren, and Jian Sun.
\newblock Deep residual learning for image recognition.
\newblock In {\em Proceedings of the IEEE conference on computer vision and pattern recognition}, pages 770--778, 2016.

\bibitem{simonyan2014very}
Karen Simonyan and Andrew Zisserman.
\newblock Very deep convolutional networks for large-scale image recognition.
\newblock {\em CoRR}, abs/1409.1556, 2014.

\bibitem{ayitey2023forex}
Michael Ayitey~Junior, Peter Appiahene, Obed Appiah, and Christopher~Ninfaakang Bombie.
\newblock Forex market forecasting using machine learning: Systematic literature review and meta-analysis.
\newblock {\em Journal of Big Data}, 10(1):9, 2023.

\bibitem{huang2019automated}
Boming Huang, Yuxiang Huan, Li~Da Xu, Lirong Zheng, and Zhuo Zou.
\newblock Automated trading systems statistical and machine learning methods and hardware implementation: a survey.
\newblock {\em Enterprise Information Systems}, 13(1):132--144, 2019.

\bibitem{lv2019empirical}
Dongdong Lv, Shuhan Yuan, Meizi Li, Yang Xiang, et~al.
\newblock An empirical study of machine learning algorithms for stock daily trading strategy.
\newblock {\em Mathematical problems in engineering}, 2019, 2019.

\bibitem{deng2016deep}
Yue Deng, Feng Bao, Youyong Kong, Zhiquan Ren, and Qionghai Dai.
\newblock Deep direct reinforcement learning for financial signal representation and trading.
\newblock {\em IEEE transactions on neural networks and learning systems}, 28(3):653--664, 2016.

\bibitem{shavandi2022multi}
Ali Shavandi and Majid Khedmati.
\newblock A multi-agent deep reinforcement learning framework for algorithmic trading in financial markets.
\newblock {\em Expert Systems with Applications}, 208:118124, 2022.

\bibitem{wu2020adaptive}
Xing Wu, Haolei Chen, Jianjia Wang, Luigi Troiano, Vincenzo Loia, and Hamido Fujita.
\newblock Adaptive stock trading strategies with deep reinforcement learning methods.
\newblock {\em Information Sciences}, 538:142--158, 2020.

\bibitem{sutton2018reinforcement}
Richard~S Sutton and Andrew~G Barto.
\newblock {\em Reinforcement learning: An introduction}.
\newblock MIT press, 2018.

\bibitem{bellemare2023distributional}
Marc~G Bellemare, Will Dabney, and Mark Rowland.
\newblock {\em Distributional reinforcement learning}.
\newblock MIT Press, 2023.

\bibitem{bellemare2017distributional}
Marc~G Bellemare, Will Dabney, and R{\'e}mi Munos.
\newblock A distributional perspective on reinforcement learning.
\newblock {\em International conference on machine learning}, pages 449--458, 2017.

\bibitem{dabney2018implicit}
Will Dabney, Georg Ostrovski, David Silver, and R{\'e}mi Munos.
\newblock Implicit quantile networks for distributional reinforcement learning.
\newblock In {\em International conference on machine learning}, pages 1096--1105. PMLR, 2018.

\bibitem{dabney2018distributional}
Will Dabney, Mark Rowland, Marc Bellemare, and R{\'e}mi Munos.
\newblock Distributional reinforcement learning with quantile regression.
\newblock In {\em Proceedings of the AAAI conference on artificial intelligence}, volume~32, 2018.

\bibitem{heche2024offline}
F{\'e}licien H{\^e}che, Oussama Barakat, Thibaut Desmettre, Tania Marx, and Stephan Robert-Nicoud.
\newblock Offline reinforcement learning in high-dimensional stochastic environments.
\newblock {\em Neural Computing and Applications}, 36(2):585--598, 2024.

\bibitem{artzner1999coherent}
Philippe Artzner, Freddy Delbaen, Jean-Marc Eber, and David Heath.
\newblock Coherent measures of risk.
\newblock {\em Mathematical finance}, 9(3):203--228, 1999.

\bibitem{sarykalin2008value}
Sergey Sarykalin, Gaia Serraino, and Stan Uryasev.
\newblock Value-at-risk vs. conditional value-at-risk in risk management and optimization.
\newblock In {\em State-of-the-art decision-making tools in the information-intensive age}, pages 270--294. Informs, 2008.

\bibitem{pinto2017robust}
Lerrel Pinto, James Davidson, Rahul Sukthankar, and Abhinav Gupta.
\newblock Robust adversarial reinforcement learning.
\newblock In {\em International Conference on Machine Learning}, pages 2817--2826. PMLR, 2017.

\bibitem{singh2020improving}
Rahul Singh, Qinsheng Zhang, and Yongxin Chen.
\newblock Improving robustness via risk averse distributional reinforcement learning.
\newblock In {\em Learning for Dynamics and Control}, pages 958--968. PMLR, 2020.

\bibitem{ma2021conservative}
Yecheng Ma, Dinesh Jayaraman, and Osbert Bastani.
\newblock Conservative offline distributional reinforcement learning.
\newblock {\em Advances in neural information processing systems}, 34:19235--19247, 2021.

\bibitem{rigter2021risk}
Marc Rigter, Bruno Lacerda, and Nick Hawes.
\newblock Risk-averse bayes-adaptive reinforcement learning.
\newblock {\em Advances in Neural Information Processing Systems}, 34:1142--1154, 2021.

\bibitem{urpi2021risk}
N{\'{u}}ria~Armengol Urp{\'{\i}}, Sebastian Curi, and Andreas Krause.
\newblock Risk-averse offline reinforcement learning.
\newblock In {\em 9th International Conference on Learning Representations, {ICLR} 2021, Virtual Event, Austria, May 3-7, 2021}. OpenReview.net, 2021.

\bibitem{chance2021introduction}
Don~M Chance and Robert Brooks.
\newblock {\em An introduction to derivatives and risk management}.
\newblock South-Western, Cengage Learning, 2021.

\bibitem{ahmed2022artificial}
Shamima Ahmed, Muneer~M Alshater, Anis El~Ammari, and Helmi Hammami.
\newblock Artificial intelligence and machine learning in finance: A bibliometric review.
\newblock {\em Research in International Business and Finance}, 61:101646, 2022.

\bibitem{levchenko2024chain}
Denis Levchenko, Efstratios Rappos, Shabnam Ataee, Biagio Nigro, and Stephan Robert-Nicoud.
\newblock Chain-structured neural architecture search for financial time series forecasting.
\newblock {\em International Journal of Data Science and Analytics}, pages 1--10, 2024.

\bibitem{nazareth2023financial}
Noella Nazareth and Yeruva Venkata~Ramana Reddy.
\newblock Financial applications of machine learning: A literature review.
\newblock {\em Expert Systems with Applications}, 219:119640, 2023.

\bibitem{jones2017predicting}
Stewart Jones, David Johnstone, and Roy Wilson.
\newblock Predicting corporate bankruptcy: An evaluation of alternative statistical frameworks.
\newblock {\em Journal of Business Finance \& Accounting}, 44(1-2):3--34, 2017.

\bibitem{lahmiri2019can}
Salim Lahmiri and Stelios Bekiros.
\newblock Can machine learning approaches predict corporate bankruptcy? evidence from a qualitative experimental design.
\newblock {\em Quantitative Finance}, 19(9):1569--1577, 2019.

\bibitem{garcia2021ai}
Olmer Garcia-Bedoya, Oscar Granados, and Jos{\'e} Cardozo~Burgos.
\newblock Ai against money laundering networks: the colombian case.
\newblock {\em Journal of Money Laundering Control}, 24(1):49--62, 2021.

\bibitem{jullum2020detecting}
Martin Jullum, Anders L{\o}land, Ragnar~Bang Huseby, Geir {\AA}nonsen, and Johannes Lorentzen.
\newblock Detecting money laundering transactions with machine learning.
\newblock {\em Journal of Money Laundering Control}, 23(1):173--186, 2020.

\bibitem{omar2017predicting}
Normah Omar, Amirah Johari, Zulaikha, and Malcolm Smith.
\newblock Predicting fraudulent financial reporting using artificial neural network.
\newblock {\em Journal of Financial Crime}, 24(2):362--387, 2017.

\bibitem{gorenc2016prediction}
Marija Gorenc~Novak and Dejan Velu{\v{s}}{\v{c}}ek.
\newblock Prediction of stock price movement based on daily high prices.
\newblock {\em Quantitative Finance}, 16(5):793--826, 2016.

\bibitem{bao2017deep}
Wei Bao, Jun Yue, and Yulei Rao.
\newblock A deep learning framework for financial time series using stacked autoencoders and long-short term memory.
\newblock {\em PloS one}, 12(7):e0180944, 2017.

\bibitem{chalvatzis2020high}
Chariton Chalvatzis and Dimitrios Hristu-Varsakelis.
\newblock High-performance stock index trading via neural networks and trees.
\newblock {\em Applied Soft Computing}, 96:106567, 2020.

\bibitem{yong2017stock}
Bang~Xiang Yong, Mohd~Rozaini Abdul~Rahim, and Ahmad~Shahidan Abdullah.
\newblock A stock market trading system using deep neural network.
\newblock In {\em Modeling, Design and Simulation of Systems: 17th Asia Simulation Conference, AsiaSim 2017}, pages 356--364. Springer, 2017.

\bibitem{breiman2001random}
Leo Breiman.
\newblock Random forests.
\newblock {\em Machine learning}, 45:5--32, 2001.

\bibitem{hochreiter1997long}
Sepp Hochreiter and J{\"u}rgen Schmidhuber.
\newblock Long short-term memory.
\newblock {\em Neural computation}, 9(8):1735--1780, 1997.

\bibitem{ghosh2022forecasting}
Pushpendu Ghosh, Ariel Neufeld, and Jajati~Keshari Sahoo.
\newblock Forecasting directional movements of stock prices for intraday trading using lstm and random forests.
\newblock {\em Finance Research Letters}, 46:102280, 2022.

\bibitem{zhang2020deep}
Zihao Zhang, Stefan Zohren, and Stephen Roberts.
\newblock Deep learning for portfolio optimization.
\newblock {\em The Journal of Financial Data Science}, 2020.

\bibitem{zhang2019extending}
Zihao Zhang, Stefan Zohren, and Stephen Roberts.
\newblock Extending deep learning models for limit order books to quantile regression.
\newblock In {\em Proceedings of Time Series Workshop of the 36th International Conference on Machine Learning, Long Beach, California, PMLR 97}, 2019.

\bibitem{marcjasz2023distributional}
Grzegorz Marcjasz, Micha{\l} Narajewski, Rafa{\l} Weron, and Florian Ziel.
\newblock Distributional neural networks for electricity price forecasting.
\newblock {\em Energy Economics}, 125:106843, 2023.

\bibitem{hambly2023recent}
Ben Hambly, Renyuan Xu, and Huining Yang.
\newblock Recent advances in reinforcement learning in finance.
\newblock {\em Mathematical Finance}, 33(3):437--503, 2023.

\bibitem{lin2021end}
Siyu Lin and Peter~A Beling.
\newblock An end-to-end optimal trade execution framework based on proximal policy optimization.
\newblock In {\em Proceedings of the Twenty-Ninth International Conference on International Joint Conferences on Artificial Intelligence}, pages 4548--4554, 2021.

\bibitem{ye2020optimal}
Zekun Ye, Weijie Deng, Shuigeng Zhou, Yi~Xu, and Jihong Guan.
\newblock Optimal trade execution based on deep deterministic policy gradient.
\newblock In {\em Database Systems for Advanced Applications, DASFAA 2020}, pages 638--654. Springer, 2020.

\bibitem{ning2021double}
Brian Ning, Franco Ho~Ting Lin, and Sebastian Jaimungal.
\newblock Double deep q-learning for optimal execution.
\newblock {\em Applied Mathematical Finance}, 28(4):361--380, 2021.

\bibitem{cao2021deep}
Jay Cao, Jacky Chen, John Hull, and Zissis Poulos.
\newblock Deep hedging of derivatives using reinforcement learning.
\newblock {\em The Journal of Financial Data Science Winter}, 3(1):10--27, 2021.

\bibitem{du2020deep}
Jiayi Du, Muyang Jin, Petter~N Kolm, Gordon Ritter, Yixuan Wang, and Bofei Zhang.
\newblock Deep reinforcement learning for option replication and hedging.
\newblock {\em The Journal of Financial Data Science}, 2(4):44--57, 2020.

\bibitem{halperin2017qlbs}
Igor Halperin.
\newblock Qlbs: Q-learner in the black-scholes (-merton) worlds.
\newblock {\em The Journal of Derivatives}, 28(1), 2020.

\bibitem{aboussalah2022value}
Amine~Mohamed Aboussalah, Ziyun Xu, and Chi-Guhn Lee.
\newblock What is the value of the cross-sectional approach to deep reinforcement learning?
\newblock {\em Quantitative Finance}, 22(6):1091--1111, 2022.

\bibitem{cong2021alphaportfolio}
Lin~William Cong, Ke~Tang, Jingyuan Wang, and Yang Zhang.
\newblock Alphaportfolio: Direct construction through reinforcement learning and interpretable ai.
\newblock {\em Social Science Research Network}, 3554486, 2020.

\bibitem{pendharkar2018trading}
Parag~C Pendharkar and Patrick Cusatis.
\newblock Trading financial indices with reinforcement learning agents.
\newblock {\em Expert Systems with Applications}, 103:1--13, 2018.

\bibitem{brim2020deep}
Andrew Brim.
\newblock Deep reinforcement learning pairs trading with a double deep q-network.
\newblock In {\em 2020 10th Annual Computing and Communication Workshop and Conference (CCWC)}, pages 0222--0227. IEEE, 2020.

\bibitem{chen2019application}
Lin Chen and Qiang Gao.
\newblock Application of deep reinforcement learning on automated stock trading.
\newblock In {\em 2019 IEEE 10th International Conference on Software Engineering and Service Science (ICSESS)}, pages 29--33. IEEE, 2019.

\bibitem{park2020intelligent}
Hyungjun Park, Min~Kyu Sim, and Dong~Gu Choi.
\newblock An intelligent financial portfolio trading strategy using deep q-learning.
\newblock {\em Expert Systems with Applications}, 158:113573, 2020.

\bibitem{sornmayura2019robust}
Sutta Sornmayura.
\newblock Robust forex trading with deep q network (dqn).
\newblock {\em ABAC Journal}, 39(1), 2019.

\bibitem{theate2021application}
Thibaut Th{\'e}ate and Damien Ernst.
\newblock An application of deep reinforcement learning to algorithmic trading.
\newblock {\em Expert Systems with Applications}, 173:114632, 2021.

\bibitem{tran2023optimizing}
Minh Tran, Duc Pham-Hi, and Marc Bui.
\newblock Optimizing automated trading systems with deep reinforcement learning.
\newblock {\em Algorithms}, 16(1):23, 2023.

\bibitem{liu2018practical}
Xiao-Yang Liu, Zhuoran Xiong, Shan Zhong, Hongyang Yang, and Anwar Walid.
\newblock Practical deep reinforcement learning approach for stock trading.
\newblock {\em NIPS 2018 Workshop on Challenges and Opportunities for AI in Financial Services}, 2018.

\bibitem{li2020application}
Yuming Li, Pin Ni, and Victor Chang.
\newblock Application of deep reinforcement learning in stock trading strategies and stock forecasting.
\newblock {\em Computing}, 102(6):1305--1322, 2020.

\bibitem{kwak2023self}
Dongkyu Kwak, Sungyoon Choi, and Woojin Chang.
\newblock Self-attention based deep direct recurrent reinforcement learning with hybrid loss for trading signal generation.
\newblock {\em Information Sciences}, 623:592--606, 2023.

\bibitem{yuan2020using}
Yuyu Yuan, Wen Wen, and Jincui Yang.
\newblock Using data augmentation based reinforcement learning for daily stock trading.
\newblock {\em Electronics}, 9(9):1384, 2020.

\bibitem{karkhanis2020ensembling}
Atharva~Abhay Karkhanis, Jayesh~Bapu Ahire, Ishana~Vikram Shinde, and Satyam Kumar.
\newblock Ensembling reinforcement learning for portfolio management.
\newblock {\em International Research Journal of Engineering and Technology (IRJET)}, 2020.

\bibitem{cao2023gamma}
Jay Cao, Jacky Chen, Soroush Farghadani, John Hull, Zissis Poulos, Zeyu Wang, and Jun Yuan.
\newblock Gamma and vega hedging using deep distributional reinforcement learning.
\newblock {\em Frontiers in Artificial Intelligence}, 6:1129370, 2023.

\bibitem{fathan2021deep}
Abderrahim Fathan and Erick Delage.
\newblock Deep reinforcement learning for optimal stopping with application in financial engineering.
\newblock {\em Les Cahiers du GERAD ISSN}, 711:2440, 2021.

\bibitem{vittori2020option}
Edoardo Vittori, Michele Trapletti, and Marcello Restelli.
\newblock Option hedging with risk averse reinforcement learning.
\newblock In {\em Proceedings of the first ACM international conference on AI in finance}, pages 1--8, 2020.

\bibitem{fei2021exponential}
Yingjie Fei, Zhuoran Yang, Yudong Chen, and Zhaoran Wang.
\newblock Exponential bellman equation and improved regret bounds for risk-sensitive reinforcement learning.
\newblock {\em Advances in neural information processing systems}, 34:20436--20446, 2021.

\bibitem{greenberg2022efficient}
Ido Greenberg, Yinlam Chow, Mohammad Ghavamzadeh, and Shie Mannor.
\newblock Efficient risk-averse reinforcement learning.
\newblock {\em Advances in Neural Information Processing Systems}, 35:32639--32652, 2022.

\bibitem{theate2023risk}
Thibaut Th{\'e}ate and Damien Ernst.
\newblock Risk-sensitive policy with distributional reinforcement learning.
\newblock {\em Algorithms}, 16(7):325, 2023.

\bibitem{tversky1992advances}
Amos Tversky and Daniel Kahneman.
\newblock Advances in prospect theory: Cumulative representation of uncertainty.
\newblock {\em Journal of Risk and uncertainty}, 5:297--323, 1992.

\bibitem{chow2014algorithms}
Yinlam Chow and Mohammad Ghavamzadeh.
\newblock Algorithms for cvar optimization in mdps.
\newblock In {\em Proceedings of the 27th International Conference on Neural Information Processing Systems}, volume~2 of {\em NIPS'14}, page 3509–3517, Cambridge, MA, USA, 2014. MIT Press.

\bibitem{chow2015risk}
Yinlam Chow, Aviv Tamar, Shie Mannor, and Marco Pavone.
\newblock Risk-sensitive and robust decision-making: a cvar optimization approach.
\newblock {\em Advances in neural information processing systems}, 28, 2015.

\bibitem{ying2022towards}
Chengyang Ying, Xinning Zhou, Hang Su, Dong Yan, Ning Chen, and Jun Zhu.
\newblock Towards safe reinforcement learning via constraining conditional value-at-risk.
\newblock {\em Thirty-First International Joint Conference on Artificial Intelligence}, 2022.

\bibitem{zhang2019deep}
Zihao Zhang, Stefan Zohren, and Stephen Roberts.
\newblock Deep reinforcement learning for trading.
\newblock {\em The Journal of Financial Data Science}, 2(2):25--40, 2020.

\bibitem{wang2022risk}
Mingfu Wang and Hyejin Ku.
\newblock Risk-sensitive policies for portfolio management.
\newblock {\em Expert Systems with Applications}, 198:116807, 2022.

\bibitem{shen2014risk}
Yun Shen, Ruihong Huang, Chang Yan, and Klaus Obermayer.
\newblock Risk-averse reinforcement learning for algorithmic trading.
\newblock In {\em 2014 IEEE conference on computational intelligence for financial engineering \& economics (CIFEr)}, pages 391--398. IEEE, 2014.

\bibitem{acerbi2002spectral}
Carlo Acerbi.
\newblock Spectral measures of risk: A coherent representation of subjective risk aversion.
\newblock {\em Journal of Banking \& Finance}, 26(7):1505--1518, 2002.

\bibitem{rockafellar2007coherent}
R~Tyrrell Rockafellar.
\newblock Coherent approaches to risk in optimization under uncertainty.
\newblock In {\em OR Tools and Applications: Glimpses of Future Technologies}, pages 38--61. Informs, 2007.

\bibitem{delbaen2002coherent}
Freddy Delbaen.
\newblock {\em Coherent risk measures on general probability spaces}, pages 1--37.
\newblock Springer, 2002.

\bibitem{rockafellar2002deviation}
R~Tyrrell Rockafellar, Stanislav~P Uryasev, and Michael Zabarankin.
\newblock Deviation measures in risk analysis and optimization.
\newblock {\em University of Florida, Department of Industrial \& Systems Engineering Working Paper}, (2002-7), 2002.

\bibitem{rockafellar2006generalized}
R~Tyrrell Rockafellar, Stan Uryasev, and Michael Zabarankin.
\newblock Generalized deviations in risk analysis.
\newblock {\em Finance and Stochastics}, 10:51--74, 2006.

\bibitem{geurts2006extremely}
Pierre Geurts, Damien Ernst, and Louis Wehenkel.
\newblock Extremely randomized trees.
\newblock {\em Machine learning}, 63:3--42, 2006.

\bibitem{pagliaro2023forecasting}
Antonio Pagliaro.
\newblock Forecasting significant stock market price changes using machine learning: Extra trees classifier leads.
\newblock {\em Electronics}, 12(21):4551, 2023.

\bibitem{mnih2015human}
Volodymyr Mnih, Koray Kavukcuoglu, David Silver, Andrei~A Rusu, Joel Veness, Marc~G Bellemare, Alex Graves, Martin Riedmiller, Andreas~K Fidjeland, Georg Ostrovski, et~al.
\newblock Human-level control through deep reinforcement learning.
\newblock {\em nature}, 518(7540):529--533, 2015.

\bibitem{van2016deep}
Hado Van~Hasselt, Arthur Guez, and David Silver.
\newblock Deep reinforcement learning with double q-learning.
\newblock In {\em Proceedings of the AAAI conference on artificial intelligence}, volume~30, 2016.

\bibitem{fujimoto2018addressing}
Scott Fujimoto, Herke Hoof, and David Meger.
\newblock Addressing function approximation error in actor-critic methods.
\newblock In {\em International conference on machine learning}, pages 1587--1596. PMLR, 2018.

\bibitem{ioffe2015batch}
Sergey Ioffe and Christian Szegedy.
\newblock Batch normalization: accelerating deep network training by reducing internal covariate shift.
\newblock In {\em Proceedings of the 32nd International Conference on International Conference on Machine Learning}, volume~37 of {\em ICML'15}, page 448–456. JMLR.org, 2015.

\bibitem{srivastava2014dropout}
Nitish Srivastava, Geoffrey Hinton, Alex Krizhevsky, Ilya Sutskever, and Ruslan Salakhutdinov.
\newblock Dropout: a simple way to prevent neural networks from overfitting.
\newblock {\em The journal of machine learning research}, 15(1):1929--1958, 2014.

\bibitem{kingma2014adam}
Diederik~P. Kingma and Jimmy Ba.
\newblock Adam: {A} method for stochastic optimization.
\newblock In {\em 3rd International Conference on Learning Representations, {ICLR} 2015, San Diego, CA, USA, May 7-9, 2015, Conference Track Proceedings}, 2015.

\bibitem{schaul2015prioritized}
Tom Schaul, John Quan, Ioannis Antonoglou, and David Silver.
\newblock Prioritized experience replay.
\newblock In {\em 4th International Conference on Learning Representations, {ICLR} 2016, San Juan, Puerto Rico, May 2-4, 2016, Conference Track Proceedings}, 2016.

\bibitem{wang2016dueling}
Ziyu Wang, Tom Schaul, Matteo Hessel, Hado Hasselt, Marc Lanctot, and Nando Freitas.
\newblock Dueling network architectures for deep reinforcement learning.
\newblock In {\em International conference on machine learning}, pages 1995--2003. PMLR, 2016.

\bibitem{huber1992robust}
Peter~J Huber.
\newblock Robust estimation of a location parameter.
\newblock In {\em Breakthroughs in statistics: Methodology and distribution}, pages 492--518. Springer, 1992.

\bibitem{heather2020european}
Patrick Heather.
\newblock European traded gas hubs: the supremacy of ttf.
\newblock {\em Oxford Institute for Energy Studies (OIES), Oxford Energy Comment}, 2020.

\bibitem{walasek2021fractional}
Rafa{\l} Walasek and Janusz Gajda.
\newblock Fractional differentiation and its use in machine learning.
\newblock {\em International Journal of Advances in Engineering Sciences and Applied Mathematics}, 13(2):270--277, 2021.

\bibitem{wold1987principal}
Svante Wold, Kim Esbensen, and Paul Geladi.
\newblock Principal component analysis.
\newblock {\em Chemometrics and intelligent laboratory systems}, 2(1-3):37--52, 1987.

\bibitem{sharpe1998sharpe}
William~F Sharpe.
\newblock The sharpe ratio.
\newblock {\em Streetwise--the Best of the Journal of Portfolio Management}, 3:169--85, 1998.

\bibitem{chen2017tutorial}
Yen-Chi Chen.
\newblock A tutorial on kernel density estimation and recent advances.
\newblock {\em Biostatistics \& Epidemiology}, 1(1):161--187, 2017.

\bibitem{al2006normalization}
Luai Al~Shalabi and Zyad Shaaban.
\newblock Normalization as a preprocessing engine for data mining and the approach of preference matrix.
\newblock In {\em 2006 International conference on dependability of computer systems}, pages 207--214. IEEE, 2006.

\bibitem{tanaka2022sliding}
Taichi Tanaka, Isao Nambu, Yoshiko Maruyama, and Yasuhiro Wada.
\newblock Sliding-window normalization to improve the performance of machine-learning models for real-time motion prediction using electromyography.
\newblock {\em Sensors}, 22(13):5005, 2022.

\bibitem{gelada2019deepmdp}
Carles Gelada, Saurabh Kumar, Jacob Buckman, Ofir Nachum, and Marc~G Bellemare.
\newblock Deepmdp: Learning continuous latent space models for representation learning.
\newblock In {\em International conference on machine learning}, pages 2170--2179. PMLR, 2019.

\bibitem{lee2020stochastic}
Alex~X Lee, Anusha Nagabandi, Pieter Abbeel, and Sergey Levine.
\newblock Stochastic latent actor-critic: Deep reinforcement learning with a latent variable model.
\newblock {\em Advances in Neural Information Processing Systems}, 33:741--752, 2020.

\bibitem{zhangrep2021}
Amy Zhang, Rowan~Thomas McAllister, Roberto Calandra, Yarin Gal, and Sergey Levine.
\newblock Learning invariant representations for reinforcement learning without reconstruction.
\newblock In {\em 9th International Conference on Learning Representations, {ICLR} 2021, Virtual Event, Austria, May 3-7, 2021}, 2021.

\bibitem{wang2000class}
Shaun~S Wang.
\newblock A class of distortion operators for pricing financial and insurance risks.
\newblock {\em Journal of risk and insurance}, pages 15--36, 2000.

\bibitem{zhou2012ensemble}
Zhi-Hua Zhou.
\newblock {\em Ensemble methods: foundations and algorithms}.
\newblock CRC press, 2012.

\end{thebibliography}

\end{document}